\documentclass{article}

\PassOptionsToPackage{numbers, compress}{natbib}

\usepackage[preprint]{neurips_2022}

\usepackage[utf8]{inputenc} %
\usepackage[T1]{fontenc}    %
\usepackage{hyperref}       %
\usepackage{url}            %
\usepackage{booktabs}       %
\usepackage{amsfonts}       %
\usepackage{nicefrac}       %
\usepackage{microtype}      %
\usepackage{xcolor}         %

\usepackage{times}
\usepackage{qtree}
\usepackage{latexsym}
\usepackage{microtype}
\usepackage{natbib}
\usepackage{booktabs}
\usepackage{appendix}
\usepackage{tikz}
\usepackage{tikz-dependency}
\usepackage{amsmath}
\usepackage{amsfonts}
\usepackage{graphicx}
\usepackage{amsthm}
\usepackage{mathtools}
\usepackage{multirow}
\usepackage{url}
\usepackage{color}
\usepackage{subfig}
\usepackage{dsfont}
\usepackage[ruled,vlined,noend]{algorithm2e}
\usepackage{wrapfig}
\usepackage{placeins}
\usepackage[noabbrev,capitalize]{cleveref} %

\crefname{equation}{equation}{equations}   %
\crefname{footnote}{footnote}{footnotes}   %
\crefname{line}{line}{lines}               %
\crefname{section}{\S}{\S\S}
\Crefname{section}{\S}{\S\S}    %
\crefformat{section}{#2\S#1#3}  %
\Crefformat{section}{#2\S#1#3}
\crefrangeformat{section}{\S\S#3#1#4--#5#2#6}
\Crefrangeformat{section}{\S\S#3#1#4--#5#2#6}

\newcommand{\exper}[1]{\textsc{#1}}
\newcommand{\expmb}[1]{\mathbf{#1}}
\newcommand{\expobj}[1]{\mathcal{L}_\text{#1}}

\newcommand{\alphabar}{\bar{\alpha}}
\newcommand{\ww}{\expmb{w}}
\newcommand{\cc}{\expmb{c}}
\newcommand{\cx}[1]{\mathbf{x}_{#1}}
\newcommand{\ptheta}{p_\theta}

\newcommand{\plm}{p_\text{lm}}
\newcommand{\qphi}{q_\phi}
\newcommand{\Lsimple}{\expobj{simple}}
\newcommand{\Lvlb}{\expobj{vlb}}
\newcommand{\Lsimpleee}{\mathcal{L}^\text{e2e}_\text{simple}}

\newcommand{\Emb}{\exper{Emb}}

\DeclareMathOperator{\Clamp}{Clamp}
\DeclareMathOperator{\argmin}{argmin}

\newcommand\errorspan[1]{\textcolor{red}{#1}}
\newcommand\corrspan[1]{\textcolor{black}{#1}}
\newcommand\greenspan[1]{\textcolor{green}{#1}}
\newcommand\tgtspan[1]{\textbf{#1}}

\usepackage{microtype}

\title{Diffusion-LM Improves Controllable Text Generation}

\author{%
  Xiang Lisa Li,  John Thickstun, Ishaan Gulrajani, Percy Liang, Tatsunori Hashimoto \\
  Computer Science Department, Stanford University\\
  \texttt{\{xlisali, jthickst,igul, thashim\}@stanford.edu}\\ \texttt{pliang@cs.stanford.edu} \\
 }

\author{%
  Xiang Lisa Li  \\
  Stanford University\\
  \texttt{xlisali@stanford.edu} \\
  \And
  John Thickstun \\
  Stanford University\\
  \texttt{jthickst@stanford.edu} \\
  \And
  Ishaan Gulrajani \\
  Stanford Univeristy \\
  \texttt{igul@stanford.edu} \\
  \And
  Percy Liang \\
  Stanford Univeristy \\
  \texttt{pliang@cs.stanford.edu} \\
  \And
  Tatsunori B. Hashimoto \\
  Stanford Univeristy \\
  \texttt{thashim@stanford.edu} \\
}

\begin{document}

\maketitle

\begin{abstract}
Controlling the behavior of language models (LMs) without re-training is a major open problem in natural language generation. While recent works have demonstrated successes on controlling simple sentence attributes (e.g., sentiment), there has been little progress on complex, fine-grained controls (e.g., syntactic structure).
To address this challenge, we develop a new \emph{non-autoregressive} language model based on \emph{continuous} diffusions that we call Diffusion-LM. 
Building upon the recent successes of diffusion models in continuous domains, Diffusion-LM iteratively denoises a sequence of Gaussian vectors into word vectors, yielding a sequence of intermediate latent variables.
The continuous, hierarchical nature of these intermediate variables enables a simple gradient-based algorithm to perform complex, controllable generation tasks.
We demonstrate successful control of Diffusion-LM for six challenging fine-grained control tasks, significantly outperforming prior work.\footnote{Code is available at \url{https://github.com/XiangLi1999/Diffusion-LM.git} }
\end{abstract}

\section{Introduction}
\label{sec:intro}

Large autoregressive language models (LMs) are capable of generating high quality text \cite{radford2019language,brown-et-al-gpt3,Chowdhery2022PaLMSL,Zhang2022OPT}, but in order to reliably deploy these LMs in real world applications, the text generation process needs to be \emph{controllable}: we need to generate text that satisfies desired requirements (e.g. topic, syntactic structure). 
A natural approach for controlling a LM would be to fine-tune the LM using supervised data of the form (control, text)
\cite{Keskar2019CTRL}. However, updating the LM parameters for each control task can be expensive and does not allow for compositions of multiple controls (e.g. generate text that is both positive sentiment \emph{and} non-toxic).
This motivates light-weight and modular plug-and-play approaches \cite{Dathathri2020Plug} that keep the LM frozen and steer the generation process using an external classifier that measures how well the generated text satisfies the control.  
But steering a frozen autoregressive LM has been shown to be difficult, and existing successes have been limited to
simple, attribute-level controls (e.g., sentiment or topic) \cite{Dathathri2020Plug,KrauseGeDi2020,yang-klein-2021-fudge}.

\begin{figure}
    \centering
    \includegraphics[width=0.8\textwidth]{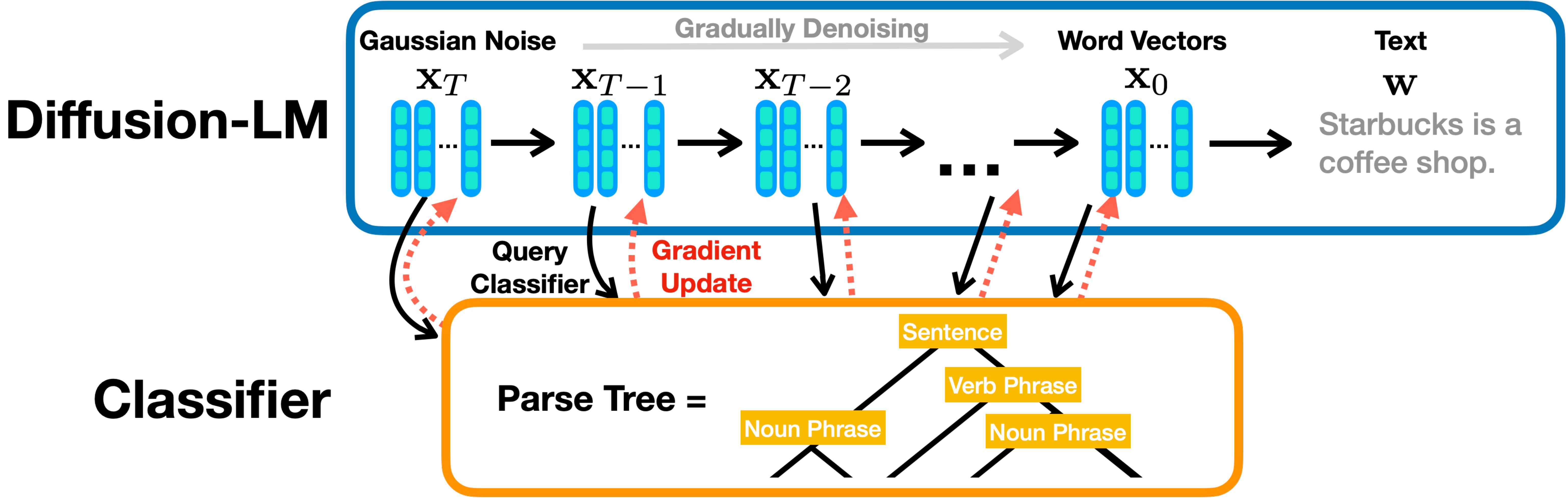}
     \caption{\label{fig:1} Diffusion-LM iteratively denoises a sequence of Gaussian vectors into word vectors, yielding a intermediate latent variables of decreasing noise level $\cx{T} \cdots \cx{0}$. For controllable generation, we iteratively perform gradient updates on these continuous latents to optimize for fluency (parametrized by Diffusion-LM) and satisfy control requirements (parametrized by a classifier). 
     \vspace{-5mm}
    }
\end{figure}

In order to tackle more complex controls,
we propose \emph{Diffusion-LM}, a new language model based on \emph{continuous} diffusions. 
Diffusion-LM starts with a sequence of Gaussian noise vectors and incrementally denoises them into vectors corresponding to words, as shown in \cref{fig:1}.
These gradual denoising steps produce a hierarchy of continuous latent representations.
We find that this hierarchical and continuous latent variable enables simple, gradient-based methods to perform complex control tasks such as constraining the parse tree of a generated sequence.

Continuous diffusion models have been extremely successful in vision and audio domains \cite{ho2020denoising,kong2020diffwave,ramesh2022hierarchical,dhariwal2021diffusion,chen2021wavegrad}, but they have not been applied to text because of the inherently discrete nature of text (\cref{sec:background}). Adapting this class of models to text requires several modifications to standard diffusions: we add an embedding step and a rounding step to the standard diffusion process, design a training objective to learn the embedding, and propose techniques to improve rounding (\cref{sec:method}).  
We control Diffusion-LM using a gradient-based method, as shown in \cref{fig:1}. This method enables us to steer the text generation process towards outputs that satisfy target structural and semantic controls. It iteratively performs gradient updates on the continuous latent variables of Diffusion-LM to balance fluency and control satisfaction (\cref{ssec:control_text}).%

To demonstrate control of Diffusion-LM, we consider six control targets ranging from fine-grained attributes (e.g., semantic content) to complex structures (e.g., parse trees). 
Our method almost doubles the success rate of previous plug-and-play methods and matches or outperforms the fine-tuning oracle on all these classifier-guided control tasks (\cref{ssec:control}). 
In addition to these individual control tasks, we show that we can successfully compose multiple classifier-guided controls to generate sentences with both desired semantic content and syntactic structure (\cref{ssec:compose}). 
Finally, we consider span-anchored controls, such as length control and infilling. Diffusion-LM allows us to perform these control tasks \emph{without} a classifier, and our Diffusion-LM significantly outperforms prior plug-and-play methods and is on-par with an autoregressive LM trained from scratch for the infilling task (\cref{ssec:infill}).

\section{Related Work}
\label{sec:related_work}

\paragraph{Diffusion Models for Text.}
Diffusion models \cite{pmlr-v37-sohl-dickstein15} have demonstrated great success in continuous data domains \cite{ho2020denoising,nichol2021improved, kong2020diffwave, Mittal2021Music}, producing images and audio that have state-of-the-art sample quality. To handle discrete data, past works have studied text diffusion models on \emph{discrete} state spaces, which defines a corruption process on discrete data (e.g., each token has some probability to be corrupted to an absorbing or random token)  \cite{austin2021structured,hoogeboom2021argmax,hoogeboom2022autoregressive}.
In this paper, we focus on \emph{continuous} diffusion models for text and to the best of our knowledge, our work is the first to explore this setting. 
In contrast to discrete diffusion LMs, our continuous diffusion LMs induce continuous latent representations, which enables efficient gradient-based methods for controllable generation.

\paragraph{Autoregressive and Non-autoregressive LMs.}
Most large pre-trained LMs are left-to-right autoregressive (e.g., GPT-3 \cite{brown-et-al-gpt3}, PaLM \cite{Chowdhery2022PaLMSL}). The fixed generation order limits the models' flexibility in many controllable generation settings, especially those that 
impose controls globally on both left and right contexts. One example is infilling, which imposes lexical control on the right contexts; another example is syntactic structure control, which controls global properties involving both left and right contexts.  
Since autoregressive LMs cannot directly condition on right contexts, prior works have developed specialized training and decoding techniques for these tasks \cite{sha-2020-gradient,donahue-etal-2020-enabling,qin-etal-2020-back}.
For example, \citet{Qin-COLD} proposed a decoding method that relaxes the discrete LM outputs to continuous variables and backpropagates gradient information from the right context.
Diffusion-LM can condition on arbitrary classifiers that look at complex, global properties of the sentence. 
There are other non-autoregressive LMs that have been developed for machine translation and speech-to-text tasks \cite{gu2018nonautoregressive,saharia2020latentalignments}.
However these methods are specialized for speech and translation settings, where the entropy over valid outputs is low,
and it has been shown that these approaches fail for language modeling \cite{ren-etal-2020-study}. 

\paragraph{Plug-and-Play Controllable Generation.}
Plug-and-play controllable generation aims to keep the LM frozen and steer its output using potential functions (e.g., classifiers). Given a probabilistic potential function that measures how well the generated text satisfies the desired control, the generated text should be optimized for both control satisfaction (measured by the potential function) and fluency (measured by LM probabilities) .  
There are several plug-and-play approaches based on autoregressive LMs: 
FUDGE \cite{yang-klein-2021-fudge} reweights the LM prediction at each token with an estimate of control satisfaction for the partial sequence;
GeDi \cite{KrauseGeDi2020} and DExperts \cite{liu-etal-2021-dexperts} reweight the LM prediction at each token with a smaller LM finetuned/trained for the control task.

The closest work to ours is PPLM \cite{Dathathri2020Plug}, which runs gradient ascent on an autoregressive LM's hidden activations to steer the next token to satisfy the control and maintain fluency. 
Because PPLM is based on autoregressive LMs, it can only generate left-to-right. This prevents PPLM from repairing and recovering errors made in previous generation steps.
Despite their success on attribute (e.g., topic) controls, we will show these plug-and-play methods for autoregressive LMs fail on more complex control tasks such as controlling syntactic structure and semantic content in \cref{ssec:control}. We demonstrate that Diffusion-LM is capable of plug-and-play controllable generation by applying classifier-guided gradient updates to the continuous sequence of latent variables induced by the Diffusion-LM.

\newcommand{\xxtheta}{f_\theta} 
\section{Problem Statement and Background}
\label{sec:background}

We first define controllable generation (\cref{ssec:control_setup}) and then review continuous diffusion models (\cref{ssec:diffusion_setup}).

\subsection{Generative Models and Controllable Generation for Text} 
\label{ssec:control_setup}
Text generation is the task of sampling $\ww$ from a trained language model $\plm (\ww)$, where $\ww = [w_1 \cdots w_n]$ is a sequence of discrete words and $\plm (\ww)$ is a probability distribution over sequences of words. Controllable text generation is the task of sampling $\ww$ from a conditional distribution $p(\ww \mid \cc)$, where $\cc$ denotes a \emph{control} variable. For syntactic control, $\cc$ can be a target syntax tree (\cref{fig:1}), while for sentiment control, $\cc$ could be a desired sentiment label. The goal of controllable generation is to generate $\ww$ that satisfies the control target $\cc$. 

Consider the plug-and-play controllable generation setting: we are given a language model $\plm(\ww)$ trained from a large amount of unlabeled text data, and for each control task, we are given a classifier $p(\cc \mid \ww)$ trained from smaller amount of labeled text data (e.g., for syntactic control, the classifier is a probabilistic parser). The goal is to utilize these two models to approximately sample from the posterior $p(\ww \mid \cc)$ via Bayes rule $p(\ww \mid \cc) \propto \plm(\ww) \cdot p(\cc \mid \ww)$. Here, $\plm(\ww)$ encourages $\ww$ to be fluent, and the $p(\cc \mid \ww)$ encourages $\ww$ to fulfill the control.

\subsection{Autoregressive Language Models}
The canonical approach to language modeling factors $\plm$ in an autoregressive left-to-right mannar, $\plm(\ww) = \plm(w_1) \prod_{i=2}^n \plm(x_i \mid x_{<i})$. In this case, text generation is reduced to the task of repeatedly predicting the next token conditioned on the partial sequence generated so far. The next token prediction $\plm(x_i \mid x_{<i})$ is often parametrized by Transformer architecture \cite{Vaswani2017Attn}. 

\subsection{Diffusion Models for Continuous Domains}
\label{ssec:diffusion_setup}

\begin{figure}
    \centering
    \includegraphics[width=0.8\textwidth]{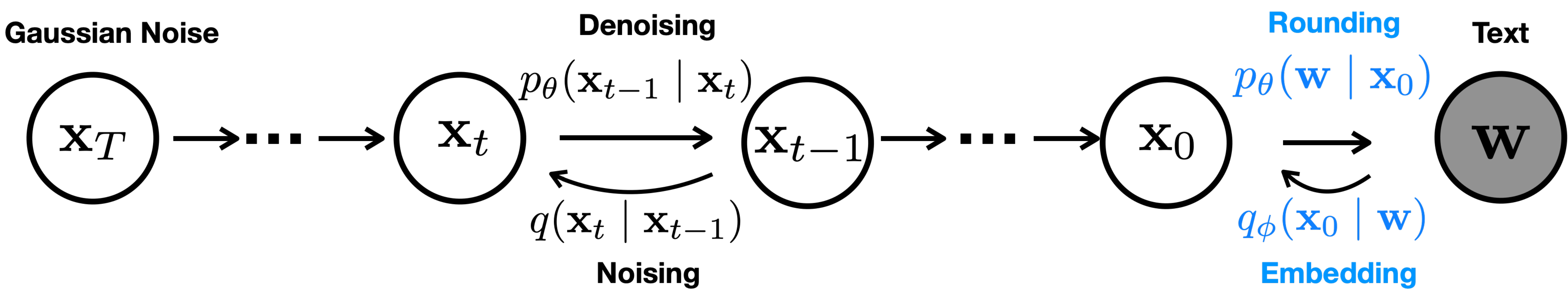}
    \caption{\label{fig:graphical_model} A graphical model representing the forward and reverse diffusion processes. In addition to the original diffusion models \cite{ho2020denoising}, we add a Markov transition between $\cx{0}$ and $\ww$, and propose the embedding \cref{ssec:embeddings} and rounding \cref{ssec:rounding} techniques. }
    \vspace{-5mm}
\end{figure}

A diffusion model \citep{ho2020denoising,nichol2021improved} is a latent variable model that
models the data $\cx{0} \in \mathbb{R}^d$ as a Markov chain $\cx{T} \dots \cx{0}$ with each variable in $\mathbb{R}^d$, and  $\cx{T}$ is a Gaussian. The diffusion model 
incrementally denoises the sequence of latent variables $\cx{T:1}$ to approximate samples from the target data distribution~(\cref{fig:graphical_model}).  The initial state $p_\theta(\cx{T}) \approx \mathcal{N} (0, \mathbf{I})$, and each denoising transition $\cx{t} \rightarrow \cx{t-1}$ is parametrized by the model $\ptheta(\cx{t-1}\mid \cx{t}) = \mathcal{N}(\cx{t-1}; \mu_\theta(\cx{t}, t), \Sigma_\theta(\cx{t}, t))$. For example, $\mu_\theta$ and $\Sigma_\theta$ may be computed by a U-Net or a Tranformer.

To train the diffusion model, we define a forward process that constructs the intermediate latent variables $\cx{1:T}$. The forward process incrementally adds Gaussian noise to data $\cx{0}$ until, at diffusion step $T$, samples $\cx{T}$ are approximately Gaussian. Each transition $\cx{t-1} \rightarrow \cx{t}$ is parametrized by $q(\cx{t} \mid \cx{t-1}) = \mathcal{N} (\cx{t} ; \sqrt{1-\beta_t} \cx{t-1}, \beta_t \mathbf{I})$, where the hyperparameter $\beta_t$ is the amount of noise added at diffusion step $t$.
This parametrization of the forward process $q$ contains no trainable parameters and allows us to define a training objective that involves generating noisy data according to a pre-defined forward process $q$ and training a model to reverse the process and reconstruct the data.

The diffusion model is trained to maximize the marginal likelihood of the data $\mathbb{E}_{\cx{0} \sim p_\text{data} }[\log p_\theta(\cx{0})]$, and the canonical objective is the variational lower bound of $\log p_\theta(\cx{0})$ \cite{pmlr-v37-sohl-dickstein15},
\begin{equation}
\Lvlb (\cx{0}) = \mathop{\mathbb{E}}_{q(\cx{1:T} | \cx{0})} \left[\log \frac{q(\cx{T} | \cx{0})}{\ptheta(\cx{T})} + \sum_{t=2}^T \log \frac{q(\cx{t-1} | \cx{0},\cx{t})} {\ptheta(\cx{t-1} | \cx{t}) } - \log \ptheta( \cx{0} | \cx{1})\right].
\label{eqn:diffusion}
\end{equation}
However, this objective can be unstable and require many optimization tricks to stabilize \cite{nichol2021improved}. To circumvent this issue, \citet{ho2020denoising} devised a simple surrogate objective that expands and reweights each KL-divergence term in $\Lvlb$ to obtain a mean-squared error loss  (derivation in \cref{app:derive}) which we will refer to as
\[\Lsimple (\cx{0}) = \sum_{t=1}^T \mathop{\mathbb{E}}_{q(\cx{t} \mid \cx{0})} || \mu_\theta(\cx{t}, t) - \hat{\mu}(\cx{t}, \cx{0}) ||^2,\]
where $\hat{\mu}(\cx{t}, \cx{0})$ is the mean of the posterior $q(\cx{t-1} | \cx{0},\cx{t})$ which is a closed from Gaussian, and $\mu_\theta(\cx{t}, t)$ is the predicted mean of $\ptheta(\cx{t-1} \mid \cx{t})$ computed by a neural network. 
While $\Lsimple$ is no longer a valid lower bound, prior work has found that it empirically made training more stable and improved sample quality\footnote{Our definition of $\Lsimple$ here uses a different parametrization from \citet{ho2020denoising}. We define our squared loss in terms of $\mu_\theta(\cx{t}, t)$ while they express it in terms of $\epsilon_\theta(\cx{t}, t)$.}.
We will make use of similar simplifications in Diffusion-LM to stabilize training and improve sample quality (\cref{ssec:embeddings}).

\vspace{-2mm}
\section{Diffusion-LM: Continuous Diffusion Language Modeling}
\vspace{-2mm}
\label{sec:method}

Constructing Diffusion-LM requires several modifications to the standard diffusion model.
First, we must define an embedding function that maps discrete text into a continuous space. To address this, we propose an end-to-end training objective for learning embeddings (\cref{ssec:embeddings}). 
Second, we require a rounding method to map vectors in embedding space back to words. To address this, we propose training and decoding time methods to facilitate rounding (\cref{ssec:rounding}).

\vspace{-2mm}
\subsection{End-to-end Training} 
\label{ssec:embeddings}

To apply a continuous diffusion model to discrete text, we define an embedding function $\Emb(w_i)$ that maps each word to a vector in $\mathbb{R}^{d}$. We define the embedding of a sequence $\ww$ of length $n$ to be: $\Emb(\ww) = [\Emb(w_1), \dots, \Emb(w_n)] \in \mathbb{R}^{nd}$. 

We propose a modification of the diffusion model training objective (Equation \ref{eqn:diffusion}) that jointly learns the diffusion model's parameters and word embeddings.
In preliminary experiments, we explored random Gaussian embeddings, as well as pre-trained word embeddings \cite{pennington-etal-2014-glove, radford2019language}. We found that these fixed embeddings are suboptimal for Diffusion-LM compared to end-to-end training\footnote{While trainable embeddings perform best on control and generation tasks, we found that fixed embeddings onto the vocabulary simplex were helpful when optimizing for held-out perplexity. We leave discussion of this approach and perplexity results to \cref{app:simplex} as the focus of this work is generation quality and not perplexity.}.

As shown in Figure \ref{fig:graphical_model}, our approach adds a Markov transition from discrete words $\ww$ to $\cx{0}$ in the forward process, parametrized by $\qphi(\cx{0} | \ww) = \mathcal{N} (\Emb(\ww), \sigma_0 I )$. In the reverse process, we add a trainable rounding step, parametrized by $\ptheta(\ww \mid \cx{0}) = \prod_{i=1}^n \ptheta(w_i \mid x_i)$, where $\ptheta(w_i \mid x_i)$ is a softmax distribution. The training objectives introduced in \cref{sec:background} now becomes
\begin{equation}
\begin{aligned}
\mathcal{L}^\text{e2e}_{\text{vlb}} (\ww) &= \mathop{\mathbb{E}}_{q_\phi(\cx{0} | \ww)}\left[\mathcal{L}_{\text{vlb}} (\cx{0}) + \log \qphi(\cx{0} | \ww) - \log \ptheta(\ww | \cx{0})]\right], \\
 \mathcal{L}^\text{e2e}_\text{simple} (\ww) &= \mathop{\mathbb{E}}_{\qphi(\cx{0:T} | \ww)} \left[\Lsimple(\cx{0}) + || \Emb(\ww) - \mu_\theta(\cx{1}, 1) ||^2  - \log \ptheta(\ww | \cx{0})\right].
\end{aligned}
\label{obj:e2e}
\end{equation}
\begin{wrapfigure}[11]{R}{0.5\textwidth}
\vspace{-10mm}
\centering
\includegraphics[width=\linewidth]{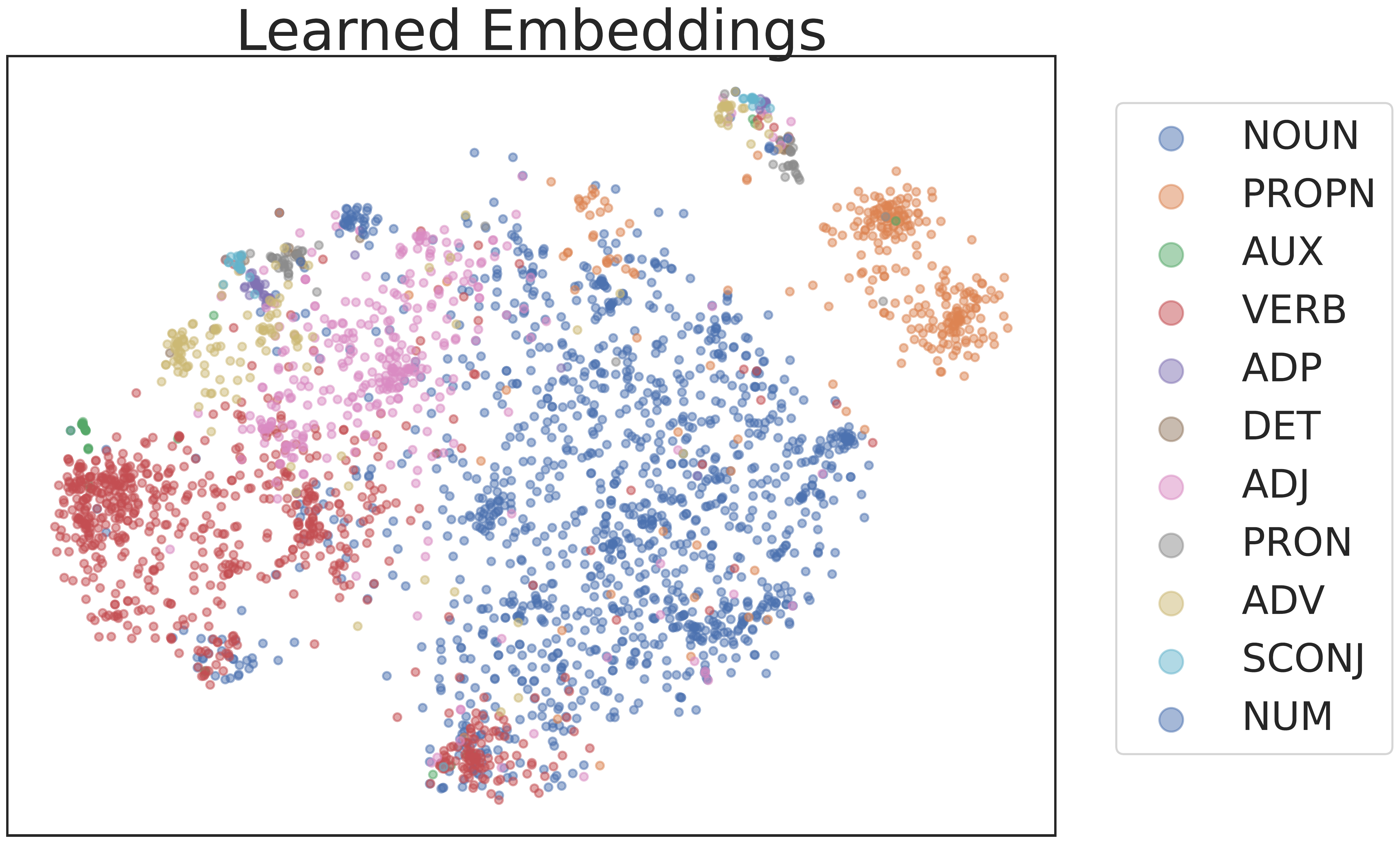}\caption{ \label{fig:emb} A t-SNE \cite{vanDerMaaten2008} plot of the learned word embeddings. Each word is colored by its POS. \looseness=-1 }
\end{wrapfigure}

We derive $ \mathcal{L}^\text{e2e}_\text{simple} (\ww)$ from $\mathcal{L}^\text{e2e}_{\text{vlb}}(\ww)$ following the simplification in \cref{ssec:diffusion_setup} and our derivation details are in \cref{app:derive}. Since we are training the embedding function, $q_\phi$ now contains trainable parameters and we use the reparametrization trick \citep{rezende2014stochastic, kingma2014auto} to backpropagate through this sampling step. 
Empirically, we find the learned embeddings cluster meaningfully: words with the same part-of-speech tags (syntactic role) tend to be clustered, as shown in \cref{fig:emb}.

\subsection{Reducing Rounding Errors}
\label{ssec:rounding}

The learned embeddings define a mapping from discrete text to the continuous $\cx{0}$. We now describe the inverse process of rounding a predicted $\cx{0}$ back to discrete text. Rounding is achieved by choosing the most probable word for each position, according to argmax $\ptheta(\ww \mid \cx{0})= \prod_{i=1}^n \ptheta(w_i \mid x_i)$. Ideally, this argmax-rounding would be sufficient to map back to discrete text, as the denoising steps should ensure that $\cx{0}$ lies exactly on the embedding of some word. However, empirically, the model fails to generate $\cx{0}$ that commits to a single word.

One explanation for this phenomenon is that the $\Lsimple(\cx{0})$ term in our objective \ref{obj:e2e} puts insufficient emphasis on modeling the structure of $\cx{0}$. Recall that we defined $\Lsimple(\cx{0}) =\sum_{t=1}^T \mathbb{E}_{\cx{t}}|| \mu_\theta(\cx{t}, t) - \hat{\mu}(\cx{t}, \cx{0}) ||^2$, where our model $\mu_\theta(\cx{t}, t)$ directly predicts the mean of  $\ptheta(\cx{t-1}\mid \cx{t})$ for each denoising step $t$. 
In this objective, the constraint that $\cx{0}$ has to commit to a single word embedding
will only appear in the terms with $t$ near $0$, and we found that this parametrization required careful tuning to force the objective to emphasize those terms (see \cref{app:ablation}).

\newcommand{\Lsimplexx}{\mathcal{L}^\text{e2e}_{\cx{0}\text{-simple}}}
Our approach re-parametrizes $\Lsimple$ to force Diffusion-LM to explicitly model $\cx{0}$ in \emph{every} term of the objective. Specifically, we derive an analogue to $\Lsimple$ which is parametrized via $\cx{0}$, 
$\Lsimplexx(\cx{0})=\sum_{t=1}^T \mathbb{E}_{\cx{t}} ||\xxtheta(\cx{t}, t) - \cx{0} ||^2$, where our model $\xxtheta(\cx{t}, t)$ predicts $\cx{0}$ directly \footnote{Predicting $\cx{0}$ and $\cx{t-1}$ is equivalent up to scaling constants as the distribution of $\cx{t-1}$ can be obtained in closed form via the forward process $\cx{t-1} = \sqrt{\alphabar} \cx{0} +\sqrt{1-\alphabar} \epsilon$, see \cref{app:derive} for further details.}. This forces the neural network to predict $\cx{0}$ in every term and we found that models trained with this objective quickly learn that $\cx{0}$ should precisely centered at a word embedding.

We described how re-parametrization can be helpful for model training, but we also found that the same intuition could be used at decoding time in a technique that we call the \emph{clamping} trick. 
In the standard generation approach for a  $\cx{0}$-parametrized model, the model denoises $\cx{t}$ to $\cx{t-1}$ by first computing an
estimate of $\cx{0}$ via $\xxtheta(\cx{t}, t)$ and then sampling $\cx{t-1}$ conditioned on this estimate: $\cx{t-1} = \sqrt{\alphabar} \xxtheta(\cx{t}, t) +\sqrt{1-\alphabar} \epsilon$, where $\alphabar_t = \prod_{s=0}^t (1-\beta_s)$ and $\epsilon \sim \mathcal{N}(0,I)$ \footnote{This follows from the marginal distribution $q(\cx{t} \mid \cx{0})$, which is a closed form Gaussian since all the Markov transitions are Gaussian.}. In the clamping trick, the model additionally maps the predicted vector $\xxtheta(\cx{t}, t)$ to its nearest word embedding sequence. Now, the sampling step becomes $\cx{t-1} = \sqrt{\alphabar}\cdot \Clamp(\xxtheta(\cx{t}, t)) +\sqrt{1-\alphabar} \epsilon$. The clamping trick forces the predicted vector to commit to a word for intermediate diffusion steps, making the vector predictions more precise and reducing rounding errors.\footnote{Intuitively, applying the clamping trick to early diffusion steps with $t$ near $T$ may be sub-optimal, because the model hasn't figured out what words to commit to. Empirically, applying clamping trick for all diffusion steps doesn't hurt the performance much. But to follow this intuition, one could also set the starting step of the clamping trick as a hyperparameter.}

\section{Decoding and Controllable Generation with Diffusion-LM}
\label{sec:decode}
Having described the Diffusion-LM, we now consider the problem of controllable text generation (\cref{ssec:control_text}) and decoding (\cref{ssec:mbr1}).

\subsection{Controllable Text Generation}
\label{ssec:control_text}
We now describe a procedure that enables plug-and-play control on Diffusion-LM. Our approach to control is inspired by the Bayesian formulation in \cref{ssec:control_setup}, but instead of performing control directly on the discrete text, we perform control on the sequence of continuous latent variables $\cx{0:T}$ defined by Diffusion-LM, and apply the rounding step to convert these latents into text.

Controlling $\cx{0:T}$ is equivalent to decoding from the posterior $p(\cx{0:T} | \cc) = \prod_{t=1}^T p(\cx{t-1} \mid \cx{t}, \cc)$, and we decompose this joint inference problem to a sequence of control problems at each diffusion step: $p(\cx{t-1} \mid \cx{t}, \cc) \propto p(\cx{t-1} \mid \cx{t}) \cdot p(\cc \mid \cx{t-1}, \cx{t})$. 
We further simplify $p(\cc \mid \cx{t-1}, \cx{t})=p(\cc \mid \cx{t-1})$ via conditional independence assumptions from prior work on controlling diffusions \cite{song2021scorebased}. Consequently, for the $t$-th step, we run gradient update on $\cx{t-1}$:
\begin{align*}
    \nabla_{\cx{t-1}} \log p(\cx{t-1} \mid \cx{t}, \cc) = \nabla_{\cx{t-1}} \log p(\cx{t-1} \mid \cx{t}) + \nabla_{\cx{t-1}}  \log  p(\cc \mid \cx{t-1}),
\end{align*}
where both $\log p(\cx{t-1} \mid \cx{t})$ and $\log p(\cc \mid \cx{t-1})$ are differentiable: the first term is parametrized by Diffusion-LM, and the second term is parametrized by a neural network classifier.

Similar to work in the image setting \cite{dhariwal2021diffusion,song2021scorebased}, we train the classifier on the diffusion latent variables and run gradient updates on the latent space $\cx{t-1}$ to steer it towards fulfilling the control. These image diffusion works take one gradient step towards $\nabla_{ \cx{t-1}} \log  p(\cc \mid \cx{t-1})$ per diffusion steps. To improve performance on text and speed up decoding,  we introduce two key modifications: fluency regularization and multiple gradient steps. \looseness=-1

To generate fluent text, we run gradient updates on a control objective with \emph{fluency regularization}: $ \lambda \log p(\cx{t-1} \mid \cx{t}) + \log p(\cc \mid \cx{t-1})$, where $\lambda$ is a hyperparameter that trades off fluency (the first term) and control (the second term).  While existing controllable generation methods for diffusions do not include the $\lambda \log p(\cx{t-1} \mid \cx{t})$ term in the objective, we found this term to be instrumental for generating fluent text. The resulting controllable generation process can be viewed as a stochastic decoding method that balances maximizing and sampling $p(\cx{t-1} \mid \cx{t}, \cc)$, much like popular text generation techniques such as nucleus sampling~\cite{Holtzman2020Nucleus} or sampling with low temperature. In order to improve the control quality, we take multiple gradient steps for each diffusion step: we run $3$ steps of the Adagrad \footnote{We tried ablations that replaced Adagrad with SGD, but we found Adagrad to be substantially less sensitive to hyperparameter tuning. } \cite{Duchi2010AdaptiveSM} update for each diffusion steps. To mitigate for the increased computation cost, we downsample the diffusion steps from 2000 to 200, which speeds up our controllable generation algorithm without hurting sample quality much.

\subsection{Minimum Bayes Risk Decoding}
\label{ssec:mbr1}
Many conditional text generation tasks require a \emph{single} high-quality output sequence, such as machine translation or sentence infilling. 
In these settings, we apply Minimum Bayes Risk (MBR) decoding \cite{kumar-byrne-2004-minimum} to aggregate a set of samples $\mathcal{S}$ drawn from the Diffusion-LM , and select the sample that achieves the minimum expected risk under a loss function $\mathcal{L}$ (e.g., negative BLEU score): $\hat{\ww} = \argmin_{\ww \in S} \sum_{\ww' \in S} \frac{1}{|S|}\mathcal{L}(\ww, \ww')$. We found that MBR decoding often returned high quality outputs, since a low quality sample would be dissimilar from the remaining samples and penalized by the loss function.

\newcommand{\Length}{Length\xspace}
\newcommand{\Spans}{Syntax Spans\xspace} 
\newcommand{\POS}{Parts-of-speech\xspace}
\newcommand{\Syntax}{Syntax Tree\xspace} 
\newcommand{\Content}{Semantic Content\xspace} 

\section{Experimental Setup}
With the above improvements on training (\cref{sec:method}) and decoding (\cref{sec:decode}), we train Diffusion-LM for two language modeling tasks. We then apply the controllable generation method to $5$ classifier-guided control tasks, and apply MBR decoding to a classifier-free control task (i.e. infilling). 

\label{sec:setup}
\subsection{Datasets and Hyperparameters} 
We train Diffusion-LM on two datasets: E2E  \cite{novikova-etal-2017-e2e} and ROCStories \cite{mostafazadeh-etal-2016-corpus}. The E2E dataset consists of 50K restaurant reviews labeled by 8 fields including food type, price, and customer rating.
The ROCStories dataset consists of 98K five-sentence stories, 
capturing a rich set of causal and temporal commonsense relations between daily events. 
This dataset is more challenging to model than E2E, because the stories contain a larger vocabulary of 11K words and more diverse semantic content.

Our Diffusion-LM is based on Transformer \cite{Vaswani2017Attn} architecture with $80$M parameters, with a sequence length $n=64$, diffusion steps $T=2000$ and a square-root noise schedule (see \cref{app:noise} for details). We treat the embedding dimension as a hyperparameter, setting $d=16$ for E2E and $d=128$ for ROCStories. See \cref{app:hyperparam} for hyperparameter details.
At decoding time, we downsample to 200 diffusion steps for E2E and maintain 2000 steps for ROCStories. Decoding Diffusion-LM for 200 steps is still 7x slower than decoding autoregressive LMs. For controllable generation, our method based on Diffusion-LM is 1.5x slower than FUDGE but 60x faster than PPLM.

\subsection{Control tasks}
\label{ssec:control_gen_setup}

\begin{table*}
    \centering
    \resizebox{0.95\linewidth}{!}{
    \begin{tabular}{lp{13cm}}
\toprule
input (\Content) & food : Japanese\\
output text & Browns Cambridge is good for Japanese food and also children friendly near The Sorrento . \\
 \midrule
input (\POS) & PROPN AUX DET ADJ NOUN NOUN VERB ADP DET NOUN ADP DET NOUN PUNCT\\
output text & Zizzi is a local coffee shop located on the outskirts of the city . \\
\midrule
input (\Syntax)  & (TOP (S (NP (*) (*) (*)) (VP (*) (NP (NP (*) (*))))))\\
output text & The Twenty Two has great food \\
\midrule
input (\Spans) & (7, 10, VP) \\ 
output text & Wildwood pub serves multicultural dishes and is ranked 3 stars\\ 
\midrule
input (\Length) & 14 \\
output text & Browns Cambridge offers Japanese food located near The Sorrento in the city centre . \\
\midrule
input (left context) & My dog loved tennis balls.\\ 
input (right context)  & My dog had stolen every one and put it under there.\\ 
output text & One day, I found all of my lost tennis balls underneath the bed. \\
\bottomrule
\end{tabular}}
\caption{\label{app:example} Example input control and output text for each control tasks.}
\end{table*}

We consider $6$ control tasks shown in \cref{app:example}:  the first 4 tasks rely on a classifier, and the last 2 tasks are classifier free\footnote{Length is classifier-free for our Diffusion-LM based methods, but other methods still require a classifier.}. 
For each control task (e.g. semantic content), we sample $200$ control targets $\cc$ (e.g., rating=5 star) from the validation splits, and we generate $50$ samples for each control target.
To evaluate the fluency of the generated text, we follow the prior works \cite{yang-klein-2021-fudge, Dathathri2020Plug} and feed the generated text to a teacher LM (i.e., a carefully fine-tuned GPT-2 model) and report the perplexity of generated text under the teacher LM. We call this metric lm-score (denoted as lm): a lower lm-score indicates better sample quality. \footnote{Prior works \cite{yang-klein-2021-fudge, Dathathri2020Plug} use GPT \cite{Radford2018ImprovingLU} as the teacher LM whereas we use a  fine-tuned GPT-2 model because our base autoregressive LM and Diffusion-LM both generate UNK tokens, which does not exist in pretrained vocabularies of GPT. }
We define success metrics for each control task as follows: 

\textbf{\Content.} Given a field (e.g., rating) and value (e.g., 5 star), generate a sentence that covers field=value, and report the success rate by exact match of `value'.

\textbf{\POS.} Given a sequence of parts-of-speech (POS) tags (e.g., \textit{Pronoun Verb Determiner Noun}), generate a sequence of words of the same length whose POS tags (under an oracle POS tagger) match the target (e.g., \textit{I ate an apple}). We quantify success via word-level exact match.

\textbf{\Syntax.} Given a target syntactic parse tree (see \cref{fig:1}),  generate text whose syntactic parse matches the given parse. To evaluate the success, we parse the generated text by an off-the-shelf parser \cite{kitaev-klein-2018-constituency}, and report F1 scores.

\textbf{\Spans.} %
Given a target (span, syntactic category) pair,  generate text whose parse tree over span $[i,j]$ matches the target syntactic category (e.g. prepositional phrase).We quantify success via the fraction of spans that match exactly.

\textbf{\Length.} Given a target length $10, \dots, 40$, generate a sequence with a length within $\pm 2$ of the target. In the case of Diffusion-LM, we treat this as a classifier-free control task. 

\textbf{Infilling.} Given a left context ($O_1$) and a right context ($O_2$) from the aNLG dataset \cite{bhagavatula2020abductive}, and the goal is to generate a sentence that logically connects $O_1$ and $O_2$. For evaluation, we report both automatic and human evaluation from the Genie leaderboard \cite{Khashabi2021GENIEAL}.

\subsection{Classifier-Guided Control Baselines}

For the first 5 control tasks, we compare our method with PPLM, FUDGE, and a fine-tuning oracle. Both PPLM and FUDGE are plug-and-play controllable generation approaches based on an autoregressive LM, which we train from scratch using the GPT-2 small architecture \cite{radford2019language}.

\textbf{PPLM\cite{Dathathri2020Plug}.}  This method runs gradient ascent on the LM activations to increase the classifier probabilities and language model probabilities, and has been successful on simple attribute control. We apply PPLM to control semantic content, but not the remaining 4 tasks which require positional information, as PPLM's classifier lacks positional information.

\textbf{FUDGE\cite{yang-klein-2021-fudge}.} For each control task, FUDGE requires a future discriminator that takes in a prefix sequence and predicts whether the complete sequence would satisfy the constraint. At decoding time, FUDGE reweights the LM prediction by the discriminator scores.

\textbf{FT.} For each control task, we fine-tune GPT-2 on (control, text) pair, yielding an \emph{oracle} conditional language model that's not plug-and-play. 
We report both the sampling (with temperature 1.0) and beam search (with beam size 4) outputs of the fine-tuned models, denoted as FT-sample and FT-search. 

\subsection{Infilling Baselines}

We compare to 3 specialized baseline methods developed in past work for the infilling task.

\textbf{DELOREAN \cite{qin-etal-2020-back}.} This method continuously relaxes the output space of a left-to-right autoregressive LM, and iteratively performs gradient updates on the continuous space to enforce fluent connection to the right contexts. This yields a continuous vector which is rounded back to text.

\textbf{COLD\cite{Qin-COLD}.} COLD specifies an energy-based model that includes fluency (from left-to-right and right-to-left LM) and coherence constraints (from lexical overlap). It samples continuous vectors from this energy-based model and round them to text.

\textbf{AR-infilling.} We train an autoregressive LM from scratch to do sentence infilling task \cite{donahue-etal-2020-enabling}. Similar to training Diffusion-LM, we train on the ROCStories dataset, but pre-process it by reordering sentences from $(O_1, O_\text{middle} , O_2)$ to $(O_1, O_2, O_\text{middle})$. At evaluation time, we feed in $O_1, O_2$, and the model generates the middle sentence.

\section{Main Results}
We train Diffusion-LMs on the E2E and ROCStories datasets.
In terms of negative log-likelihood (NLL, lower is better), we find that the variational upper bound of Diffusion-LM NLL \footnote{Exact log-likelihoods are intractable for Diffusion-LM, so we report the lower bound $\mathcal{L}^\text{e2e}_{\text{vlb}}$.} underperforms the equivalent autoregressive Transformer model (2.28 vs. 1.77 for E2E, 3.88 vs 3.05 for ROCStories) although scaling up model and dataset size partially bridges the gap (3.88 $\xrightarrow{}$ 3.10 on ROCStories).
Our best log-likelihoods required several modifications from \cref{sec:method}; we explain these and give detailed log-likelihood results in \cref{app:simplex}.
Despite worse likelihoods, controllable generation based on our Diffusion-LM results in significantly better outputs than systems based on autoregressive LMs, as we will show in \cref{ssec:control},\cref{ssec:compose}, and \cref{ssec:infill}

\newcommand{\fluency}{lm $\downarrow$}
\newcommand{\success}{ctrl $\uparrow$}

\subsection{Classifier-Guided Controllable Text Generation Results}
\label{ssec:control}
\begin{table*}
\centering
\resizebox{1.0\columnwidth}{!}{
\begin{tabular}{lcc|cc|cc|cc|cc}
\toprule
          & \multicolumn{2}{c|}{\Content} & \multicolumn{2}{c|}{\POS} & \multicolumn{2}{c|}{\Syntax} & \multicolumn{2}{c|}{\Spans} & \multicolumn{2}{c}{\Length}  \\
& \success        & \fluency       & \success     & \fluency    & \success     & \fluency     & \success     & \fluency    &  \success     & \fluency    \\ \hline
PPLM      &    9.9   & 5.32  &  - & - & - & - & - & - & - &  - \\
FUDGE     & 69.9 & 2.83  & 27.0  & 7.96     & 17.9   & \textbf{3.39} & 54.2 & 4.03 & 46.9  & 3.11   \\
Diffusion-LM & \textbf{81.2}  & \textbf{2.55}  &\textbf{90.0} & \textbf{5.16 }   & \textbf{86.0}    & 3.71 & \textbf{93.8} & \textbf{2.53}  & \textbf{99.9} & \textbf{2.16}   \\
\midrule
 FT-sample  & 72.5  & 2.87  & 89.5 & 4.72    & 64.8   & 5.72 & 26.3 & 2.88  & 98.1 & 3.84\\
FT-search  & 89.9  & 1.78  & 93.0 & 3.31    & 76.4   & 3.24 & 54.4 & 2.19  & 100.0 & 1.83\\
\bottomrule
\end{tabular}
}
\caption{\label{tab:control} Diffusion-LM achieves high success rate (\success) and good fluency (\fluency) across all 5 control tasks, outperforming the PPLM and FUDGE baselines. Our method even outperforms the fine-tuning oracle (FT) on controlling syntactic parse trees and spans.}
\end{table*}
As shown in \cref{tab:control}, Diffusion-LM achieves high success and fluency across all classifier-guided control tasks. It significantly outperforms the PPLM and FUDGE baselines across all 5 tasks. Surprisingly, our method outperforms the fine-tuning oracle on controlling syntactic parse trees and spans, while achieving similar performance on the remaining 3 tasks.  

Controlling syntactic parse trees and spans are challenging tasks for fine-tuning, because conditioning on the parse tree requires reasoning about the nested structure of the parse tree, and conditioning on spans requires lookahead planning to ensure the right constituent appears at the target position.

We observe that PPLM fails in semantic content controls and conjecture that this is because PPLM is designed to control coarse-grained attributes,
and may not be useful for more targeted tasks such as enforcing that a restaurant review contains a reference to Starbucks.

FUDGE performs well on semantic content control but does not perform well on the remaining four tasks. Controlling a structured output (\POS and \Syntax) is hard for FUDGE because making one mistake anywhere in the prefix makes the discriminator assign low probabilities to all continuations. In other control tasks that require planning (\Length and \Spans), the future discriminator is difficult to train, as it must implicitly perform lookahead planning. 

The non-autoregressive nature of our Diffusion-LM allows it to easily solve all the tasks that require precise future planning (\Spans and \Length). We believe that it works well for complex controls that involve global structures (\POS, \Syntax) because the coarse-to-fine representations allow the classifier to exert control on the entire sequence (near $t=T$) as well as on individual tokens (near $t=0$).

\paragraph{Qualitative Results.} 
\Cref{app:qualitative-syntax} shows samples of \Syntax control. Our method and fine-tuning both provide fluent sentences that mostly satisfy controls, whereas FUDGE deviates from the constraints after the first few words. 
One key difference between our method and fine-tuning is that Diffusion-LM is able to correct for a failed span and have suffix spans match the target. In the first example, the generated span (``Family friendly Indian food'') is wrong because it contains 1 more word than the target. Fortunately, this error doesn't propagate to later spans, since Diffusion-LM adjusts by dropping the conjunction. Analogously, in the second example, the FT model generates a failed span (``The Mill'') that contains 1 fewer word. However, the FT model fails to adjust in the suffix, leading to many misaligned errors in the suffix. 
\begin{table*}
    \centering
    \resizebox{0.95\linewidth}{!}{
    \begin{tabular}{lp{14cm}}
    \toprule
    Syntactic Parse & ( S ( S ( NP * ) ( VP * ( NP ( NP * * ) ( VP * \tgtspan{( NP ( ADJP * * ) * )} ) ) ) ) * ( S ( NP * * * ) ( VP * ( ADJP ( ADJP * ) ) ) ) )\\
    \midrule
FUDGE &  \errorspan{Zizzi is a cheap restaurant . \textbf{[incomplete]}} \\
Diffusion-LM &   Zizzi is a pub providing \errorspan{\textbf{family friendly Indian food}} Its customer rating is low \\
FT &  Cocum is a Pub serving \corrspan{\textbf{moderately priced meals}} and the customer rating is high \\
\toprule
     Syntactic Parse  &  ( S ( S ( VP * ( PP * ( NP * * ) ) ) ) * \tgtspan{( NP * * * )} ( VP * ( NP ( NP * * ) ( SBAR ( WHNP * ) ( S ( VP * ( NP * * ) ) ) ) ) ) * ) \\
\midrule
FUDGE &   \errorspan{In the city near The Portland Arms is a coffee and fast food place named The Cricketers which is not family - friendly with a customer rating of 5 out of 5 .} \\
Diffusion-LM &  Located on the riverside , \textbf{\corrspan{The Rice Boat}} is a restaurant that serves Indian food . \\
FT &  Located near The Sorrento, \errorspan{\textbf{The Mill} is a pub that serves Indian cuisine.} \\
    \bottomrule
    \end{tabular}}
\caption{\label{app:qualitative-syntax}  Qualitative examples from the \Syntax control. The syntactic parse tree is linearized by nested brackets representing the constituents, and we use the standard PTB syntactic categories. Tokens within each span are represented as * . We color failing spans \errorspan{red} and \textbf{bold} the spans of interest that we discuss in \cref{ssec:control}. }
\end{table*}

\subsection{Composition of Controls}
\label{ssec:compose}
\begin{table*}
    \centering
    \resizebox{0.85\linewidth}{!}{
\begin{tabular}{lccc|ccc}
\toprule

          & \multicolumn{3}{c|}{\Content + \Syntax} & \multicolumn{3}{c}{\Content + \POS }  \\
& semantic \success     & syntax \success      & \fluency       & semantic \success    & POS \success     & \fluency    \\ \hline
FUDGE           &  61.7     & 15.4   & 3.52 & 64.5 & 24.1 & 3.52\\
Diffusion-LM    &  \textbf{69.8}     & \textbf{74.8}  & 5.92 &  \textbf{63.7} & \textbf{69.1} & 3.46\\

\midrule

FT-PoE              &  61.7     & 29.2  & \textbf{2.77} &  29.4 & 10.5 & \textbf{2.97} \\
\bottomrule
\end{tabular}
 }
\caption{\label{tab:compose} In this experiment, we compose semantic control and syntactic control: Diffusion-LM achieves higher success rate (\success) at some cost of fluency (\fluency). Our method outperforms both FUDGE and FT-PoE (product of experts of two fine-tuned models) on control success rate, especially for the structured syntactic controls (i.e.  syntactic parse tree and POS).
}
\end{table*}

One unique capability of plug-and-play controllable generation is its modularity. Given
classifiers for multiple independent tasks, gradient guided control makes it simple
to generate from the intersection of multiple controls by taking gradients on the sum of the classifier log-probabilities.

We evaluate this setting on the combination of \Content + \Syntax control and \Content + \POS control.
As shown in \cref{tab:compose}, our Diffusion-LM achieves a high success rate for both of the two components, whereas FUDGE gives up on the more global syntactic control. This is expected because FUDGE fails to control syntax on its own.

Fine-tuned models are good at POS and semantic content control individually but do not compose these two controls well by product of experts (PoE), leading to a large drop in success rates for both constraints.

\subsection{Infilling Results}
\label{ssec:infill}

\begin{table}
    \centering
    \resizebox{0.8\columnwidth}{!}{
\begin{tabular}{l|cccc|cccc}
\toprule
  & \multicolumn{4}{c|}{Automatic Eval} & \multicolumn{4}{c}{Human Eval} \\
  & BLEU-4 $\uparrow$ & ROUGE-L $\uparrow$ & CIDEr $\uparrow$ & BERTScore $\uparrow$ &  \\ \hline
Left-only & 0.9 & 16.3 & 3.5 & 38.5 & n/a\\
DELOREAN & 1.6  & 19.1 & 7.9  & 41.7 & n/a\\ 
COLD & 1.8 & 19.5 & 10.7 & 42.7  & n/a\\
Diffusion & \textbf{7.1} & \textbf{28.3} & \textbf{30.7} & \textbf{89.0}  & $\textbf{0.37}^{+0.03}_{-0.02}$\\
\midrule
AR & 6.7 & 27.0 & 26.9 & \textbf{89.0}  & $\textbf{0.39}^{+0.02}_{-0.03}$\\
\bottomrule
\end{tabular}}
\vspace{2mm}
\caption{\label{tab:infill} For sentence infilling, Diffusion-LM significantly outperforms prior work COLD \cite{Qin-COLD} and Delorean  \cite{qin-etal-2020-back} (numbers taken from paper), and matches the performance of an autoregressive LM (AR) trained from scratch to do infilling.   }
\end{table}

As shown in \cref{tab:infill}, our diffusion LM significantly outperforms continuous relaxation based methods for infilling (COLD and DELOREAN). Moreover, our method achieves comparable performance to fine-tuning a specialized model for this task. Our method has slightly better automatic evaluation scores and the human evaluation found no statistically significant improvement for either method. These results suggest that Diffusion LM can solve many types of controllable generation tasks that depend on generation order or lexical constraints (such as infilling) without specialized training.

\subsection{Ablation Studies}
\label{sec:ablation}

\begin{wrapfigure}[12]{R}{0.55\textwidth}

 \vspace{-10pt}
    \centering
    \includegraphics[width=0.25\textwidth]{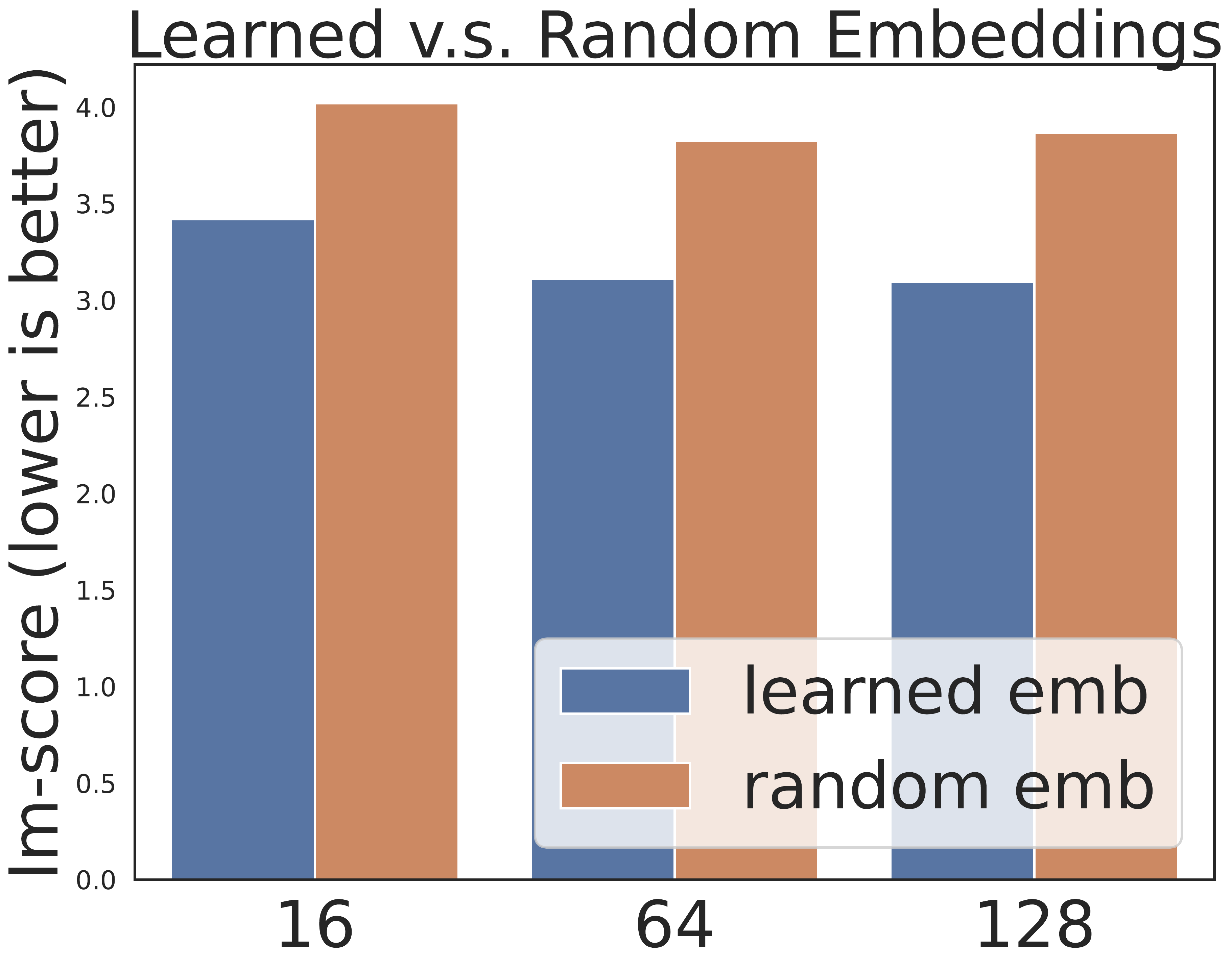}
    \includegraphics[width=0.25\textwidth]{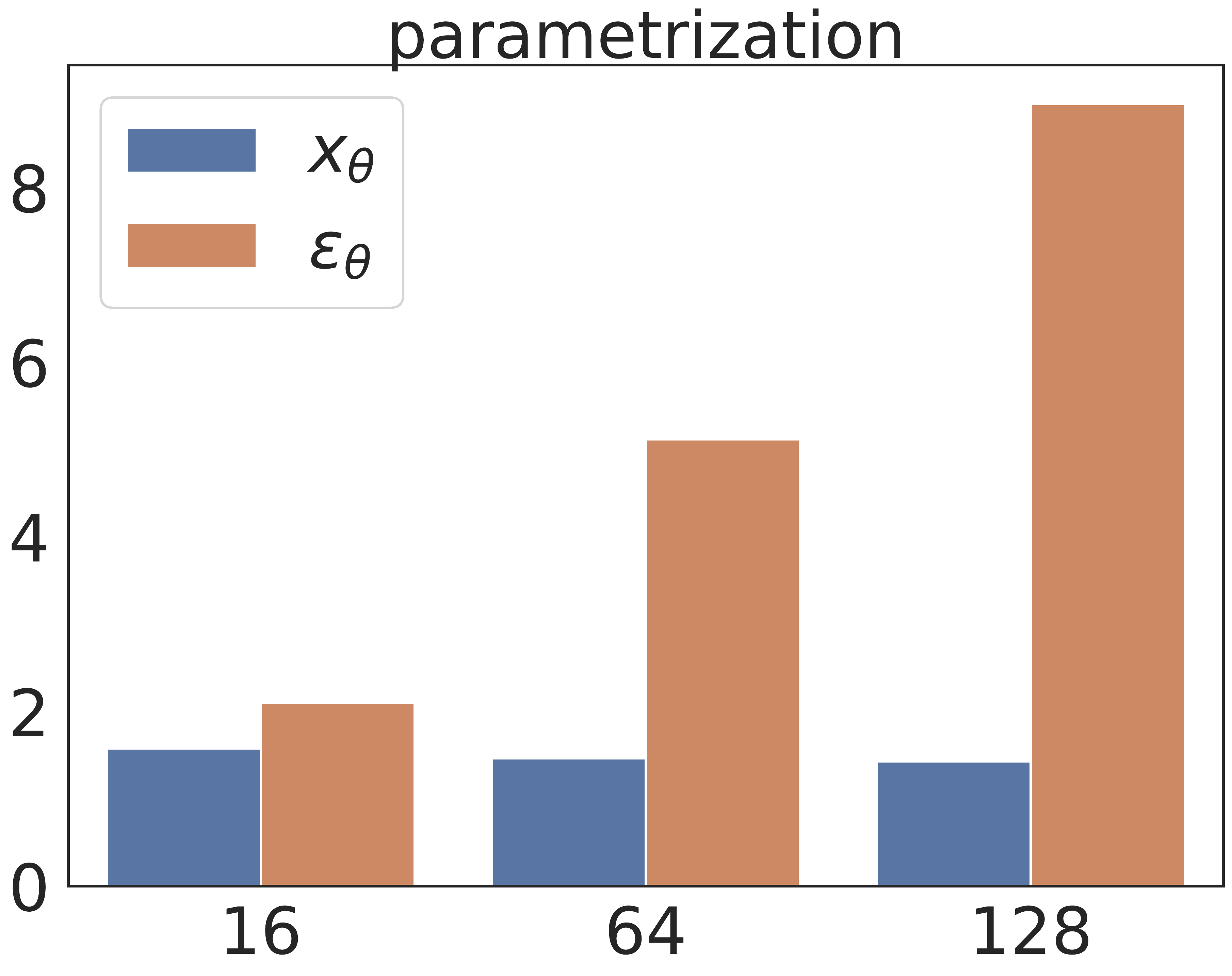}
\caption{\label{fig:ablation}
We measure the impact of our proposed design choices through lm-score. We find both learned embeddings and reparametrization substantially improves sample quality.
}
\end{wrapfigure}

We verify the importance of our proposed design choices in \cref{sec:method} through two ablation studies. We measure the sample quality of Diffusion-LM using the lm-score on 500 samples~\cref{ssec:control_gen_setup}.

\textbf{Learned v.s. Random Embeddings (\cref{ssec:embeddings}).} 
Learned embeddings outperform random embeddings on the ROCStories, which is a harder language modeling task. The same trend holds for the E2E dataset but with a smaller margin.

\textbf{Objective Parametrization (\cref{ssec:rounding}).}
We propose to let the diffusion model predict $\cx{0}$ directly. Here, we compare this with standard parametrization in image generation which parametrizes by the noise term $\epsilon$. \cref{fig:ablation} (right) shows that parametrizing by $\cx{0}$ consistently attains good performance across dimensions, whereas parametrizing by $\epsilon$ works fine for small dimensions, but quickly collapses for larger dimensions.

\section{Conclusion and Limitations}
We proposed Diffusion-LM, a novel and controllable language model based on continuous diffusions, which enables new forms of complex fine-grained control tasks.
We demonstrate Diffusion-LM’s success in 6 fine-grained control tasks: our method almost doubles the control success rate of prior methods and is competitive with  baseline fine-tuning methods that require additional training.

We find the complex controls enabled by Diffusion-LM to be compelling, and we are excited by how Diffusion-LM is a  substantial departure from the current paradigm of discrete autoregressive generation. As with any new technologies, there are drawbacks to the Diffusion-LMs that we constructed: (1) it has higher perplexity; (2) decoding is substantially slower; %
and (3) training converges more slowly. We believe that with more follow-up work and optimization,  many of these issues can be addressed, and this approach will turn out to be a compelling way to do controllable generation at scale.%

\begin{ack}
We thank Yang Song, Jason Eisner, Tianyi Zhang, Rohan Taori, Xuechen Li, Niladri Chatterji, and the members of p-lambda group for early discussions and feedbacks. We gratefully acknowledge the support of a PECASE award. 
Xiang Lisa Li is supported by a Stanford Graduate Fellowship. 
\end{ack}

\bibliography{anthology}

\clearpage

 \appendix

\section{Diffusion Noise Schedule}
\newcommand{\sqrtt}{\textit{sqrt}\xspace} 
\newcommand{\cosine}{\textit{cosine}\xspace}
\newcommand{\linear}{\textit{linear}\xspace}

\label{app:noise}
\begin{wrapfigure}[11]{R}{0.25\textwidth}
\centering
\vspace{-10pt}
\includegraphics[width=0.9\linewidth]{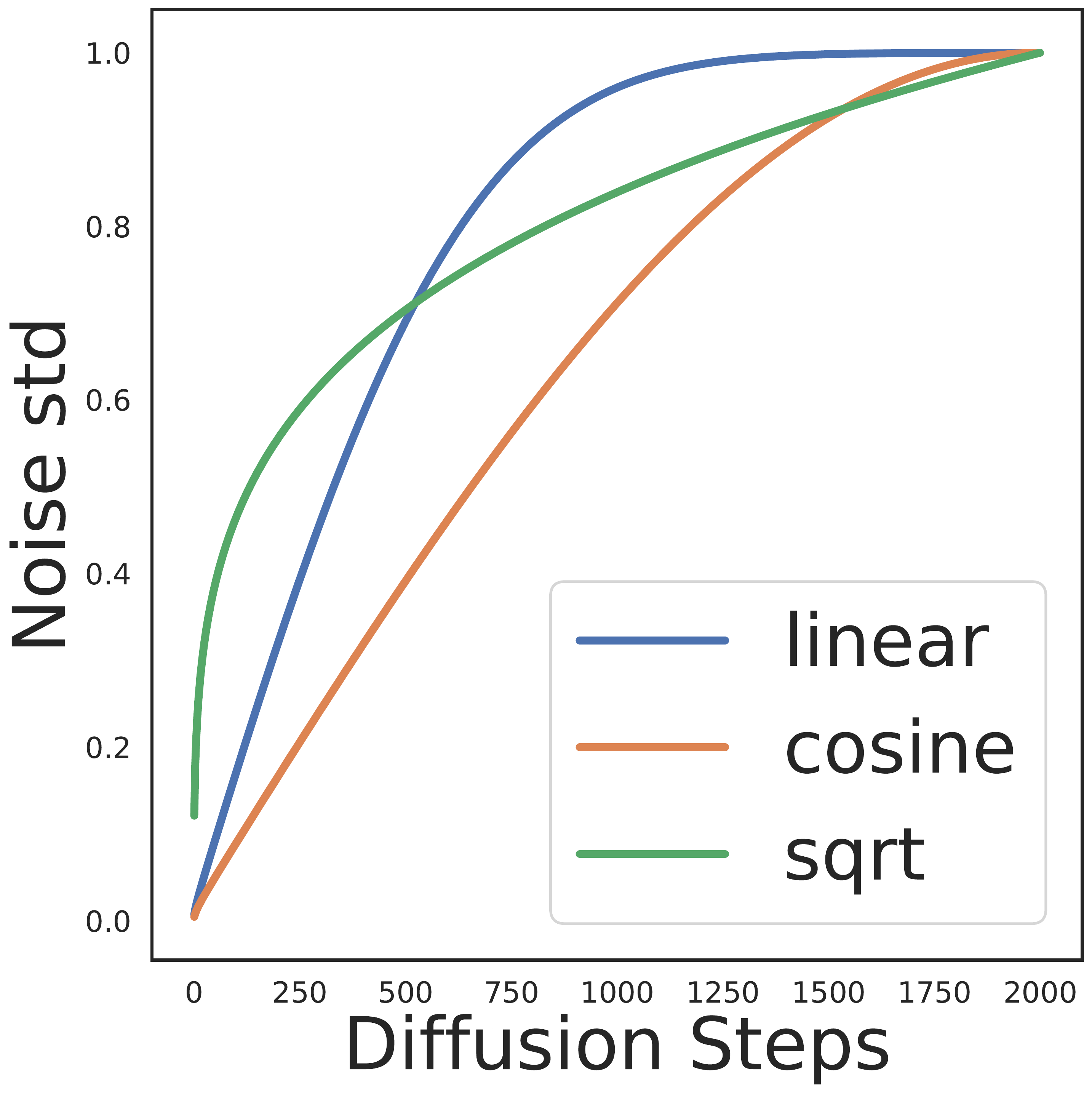}
\vspace{-3pt}
\caption{Visualizing the noise schedule $\sqrt{1-\alphabar_t}$. \label{fig:noise}  }
\vspace{-10pt}
\end{wrapfigure}

Because a diffusion model shares parameters for all diffusion steps, the  noise schedule (parametrized by $\alphabar_{1:T}$) is an important hyperparameter that determines how much weight we assign to each denoising problem. 
We find that standard noise schedules for continuous diffusions are not robust for text data. We hypothesize that the discrete nature of text and the rounding step make the model insensitive to noise near $t= 0$. Concretely, adding small amount of Gaussian noise to a word embedding is unlikely to change its nearest neighbor in the embedding space, making denoising an easy task near $t=0$.

To address this, we introduce a new \sqrtt noise schedule that is better suited for text, shown in \cref{fig:noise} defined by
$\alphabar_t = 1-\sqrt{t/T + s}$, where $s$ is a small constant that corresponds to the starting noise level\footnote{We set $s=$1e-4, and $T=2000$, which sets the initial standard deviation to $0.1$. }. Compared to standard \linear and \cosine schedules, our \sqrtt schedule starts with a higher noise level and increase noise rapidly for the first 50 steps. Then \sqrtt slows down injecting noise to avoid spending much steps in the high-noise problems, which may be too difficult to solve well.

\section{Hyperparameters} 
\label{app:hyperparam} 
\paragraph{Diffusion-LM hyperparameters.}
The hyperparameters that are specific to Diffusion-LM include the number of diffusion steps, the architecture of the Diffusion-LM, the embedding dimension, and the noise schedule, . We set the diffusion steps to be $2000$, the architecture to be  BERT-base \cite{Devlin2019BERTPO}, and the sequence length to be $64$. 
For the embedding dimensions, we select from $d \in \{16, 64, 128, 256\}$ and select $d=16$ for the E2E dataset and $d=128$ for ROCStories. 
For the noise schedule, we design the \sqrtt schedule (\cref{app:noise}) that is more robust to different parametrizations and embedding dimensions as shown in \cref{app:ablation}. However, once we picked the $\cx{0}$-parametrization (\cref{ssec:rounding}) the advantage of \sqrtt schedule is not salient. 

\paragraph{Training hyperparameters.}
We train Diffusion-LMs using AdamW optimizer and a linearly decay learning rate starting at 1e-4, dropout of 0.1, batch size of 64, and the total number of training iteration is 200K for E2E dataset, and 800K for ROCStories dataset. Our Diffusion-LMs are trained on a single GPU: NVIDIA RTX A5000, NVIDIA GeForce RTX 3090, or NVIDIA A100. It takes approximately 5 hours to train for 200K iterations on a single A100 GPU.  

To stablize the training under $\Lvlb^{\text{e2e}}$ objective, we find that we need to set gradient clipping to 1.0 and apply importance sampling to reweight each term in $\Lvlb$ \cite{nichol2021improved}. Both tricks are not necessary for $\Lsimple^{\text{e2e}}$ objective.  

\paragraph{Controllable Generation hyperparameters.} To achieve controllable generation, we run gradient update on the continuous latents of Diffusion-LM. 
We use the AdaGrad optimizer \cite{Duchi2010AdaptiveSM} to update the latent variables, and we tune the learning rate, $\text{lr} \in \{0.05, 0.1, 0.15, 0.2\}$ and the trade-off parameter $\lambda \in \{0.1, 0.01, 0.001, 0.0005\}$. Different plug-and-play controllable generation approaches tradeoff between fluency and control by tunning different hyperparameters: PPLM uses the number of gradient updates per token, denoted as $k$, and we tune $k \in \{10, 30\}$. FUDGE uses the tradeoff parameter $\lambda_{\text{FUDGE}}$ and we tune this $\lambda_{\text{FUDGE}} \in \{16, 8, 4, 2 \}$. \cref{tab:hyper} contains all the selected hyperparameter for each control tasks. Both PPLM and FUDGE has additional hyperparameters and we follow the instruction from the original paper to set those. For PPLM, we set the learning rate to be 0.04 and KL-scale to be 0.01. For FUDGE, we set precondition top-K to be 200, post top-K to be 10. 

\newcommand{\tradeoff}{tradeoff}
\begin{table*}
\centering
\resizebox{0.9\columnwidth}{!}{
\begin{tabular}{lcc|cc|cc|cc|cc}
\toprule
          & \multicolumn{2}{c|}{Semantic} & \multicolumn{2}{c|}{Parts-of-speech} & \multicolumn{2}{c|}{Syntax Tree} & \multicolumn{2}{c|}{Syntax Spans} & \multicolumn{2}{c}{Length}  \\
& \tradeoff        & lr       & \tradeoff     & lr    & \tradeoff     & lr     & \tradeoff     & lr    &  \tradeoff     & lr    \\ \hline
PPLM      &  30  & 0.04  &  - & - & - & - & - & - & - &  - \\
FUDGE     & 8.0  & -  &20.0 & - & 20.0 & - & 20.0 & - & 2.0 & -  \\
Diffusion-LM & 0.01 & 0.1 & 0.0005 & 0.05 & 0.0005 & 0.2 &  0.1 & 0.15 & 0.01 & 0.1 \\
\bottomrule
\end{tabular}
}
\caption{\label{tab:hyper} Hyperparameters for controllable generation methods. }
\end{table*}

\section{Decoding Speed}
Sampling from Diffusion-LMs requires iterating through the 2000 diffusion steps, yielding $O(2000)$ $\xxtheta$ model calls. In contrast, sampling from autoregressive LMs takes $O(n)$ where $n$ is the sequence length. Therefore, decoding Diffusion-LM is slower than decoding autoregressive LMs in short and medium-length sequence regimes. Concretely, it takes around 1 minute to decode 50 sequence of length 64.

To speed up decoding, we tried skipping steps in the generative diffusion process and downsample 2000 steps to 200 steps. Concretely, we set $T=200$ and downsample the noise schedule $\alphabar_{t} = \alphabar_{10t}$, which is equivalent to setting each unit transition as the transition $\cx{t} \rightarrow \cx{t+10}$. We decode Diffusion-LM using this new noise schedule and discretization. We find that this naive approach doesn't hurt sample quality for simple language modeling tasks like E2E, but it hurts sample quality for harder language modeling tasks like ROCStories. 

For plug-and-play controllable generation tasks, extant approaches are even slower. 
PPLM takes around 80 minutes to generate 50 samples (without batching), because it needs to run 30 gradient updates for each token. 
FUDGE takes 50 seconds to generate 50 samples (with batching), because it needs to call the lightweight classifier for each partial sequence, requiring 200 classifier calls for each token, yielding $100 \times$ sequence length calls. We can batch the classifier calls, but it sometimes limits batching across samples due to limited GPU memory. 
Our Diffusion-LM takes around 80 seconds to generate 50 samples (with batching). Our method downsamples the number of diffusion steps to 200, and it takes 3 classifier calls per diffusion step, yielding 600 model calls in total. 

\section{Classifiers for Classifier-Guided Controls}
\textbf{Semantic Content.} We train an autoregressive LM (GPT-2 small architecture) to predict the (field, value) pair conditioned on text. To parametrize $\log p(\cc \mid \cx{t})$, we compute the logprob of ``value'' per token.

\textbf{Parts-of-speech.} The classifier is parametrized by a parts-of-speech tagger, which estimates the probability of the target POS sequence conditioned on the latent variables. This tagger uses a BERT-base architecture: the input is the concantenated word embedding, and output a softmax distribution over all POS tags for each input word. $\log p(\cc \mid  \cx{t})$ is the sum of POS log-probs for each word in the sequence.  

\textbf{Syntax Tree.}  We train a Transformer-based constituency parser \cite{kitaev-klein-2018-constituency}. Our parser makes locally normalized prediction for each span, predicting either ``not a constituent'', or a label for the constituent (e.g., Noun Phrase). 
$\log p(\cc \mid  \cx{t})$ is the sum of log-probs for each labeled and non-constituency span in the sequence.

\textbf{Syntax Span.} We use the same parser trained for the syntax tree. $\log p(\cc \mid  \cx{t})$ is the log-probability that the target span is annotated with the target label.  

\section{End-to-end Objective Derivations}
\label{app:derive}
For continuous diffusion models (\cref{ssec:diffusion_setup}), $\Lsimple$ is derived from the canonical objective $\Lvlb$ by reweighting each term. 
The first $T$ terms in $\Lvlb$ are all KL divergence between two Gaussian distributions, which has a closed form solution. Take the $t$-th term for example:
\begin{equation}
\mathop{\mathbb{E}}_{q(\cx{1:T} | \cx{0})} \left[ \log \frac{q(\cx{t-1} | \cx{0},\cx{t})} {\ptheta(\cx{t-1} | \cx{t}) }\right] = \mathop{\mathbb{E}}_{ q(\cx{1:T} | \cx{0})}  \left[ \frac{1}{2\sigma_t^2}  || \mu_\theta(\cx{t}, t) - \hat{\mu}(\cx{t}, \cx{0}) ||^2\right] + C, 
\label{app:eqn:simple}
\end{equation}
where $C$ is a constant, $\hat{\mu}$ is the mean of the posterior  $q(\cx{t-1} | \cx{0},\cx{t})$, and $\mu_\theta$ is the mean of $\ptheta(\cx{t-1} \mid \cx{t})$ predicted by the diffusion model. Intuitively, this simplification matches the predicted mean of $\cx{t-1}$ to its true posterior mean. The simplification involves removing the constant $C$ and the scaling factor $\frac{1}{2\sigma_t^2}$, yielding one term in $\Lsimple$:  $\mathop{\mathbb{E}}_{ q(\cx{1:T} | \cx{0})}  \left[  || \mu_\theta(\cx{t}, t) - \hat{\mu}(\cx{t}, \cx{0}) ||^2\right]$. 

To apply continuous diffusion to model discrete text, we design Diffusion-LM (\cref{ssec:embeddings}) and propose a new end-to-end training objective (\cref{obj:e2e}) that learns the diffusion model and the embedding parameters jointly. The $\Lvlb^{\text{e2e}}$ can be written out as 
\begin{align*}
    \mathcal{L}^\text{e2e}_{\text{vlb}} (\ww) &= \mathop{\mathbb{E}}_{q_\phi(\cx{0} | \ww)}\left[\mathcal{L}_{\text{vlb}} (\cx{0}) + \log \qphi(\cx{0} | \ww) - \log \ptheta(\ww | \cx{0})]\right] \\
    &= \mathop{\mathbb{E}}_{q_\phi(\cx{0:T} | \ww)}  \left[\underbrace{\log \frac{q(\cx{T} | \cx{0})}{\ptheta(\cx{T})}}_{L_T} + \sum_{t=2}^T \underbrace{\log \frac{q(\cx{t-1} | \cx{0},\cx{t})} {\ptheta(\cx{t-1} | \cx{t}) }}_{L_{t-1}} - \underbrace{\frac{\log \qphi(\cx{0} | \ww)}{\log \ptheta( \cx{0} | \cx{1})}}_{L_{0}}     - \underbrace{\log \ptheta(\ww | \cx{0})}_{L_\text{round}}\right]  \\
\end{align*}
We apply the same simplification which transforms $\Lvlb \rightarrow \Lsimple$ to transform  $\mathcal{L}^\text{e2e}_{\text{vlb}} \rightarrow \Lsimpleee$: 
\begin{align*}
    \mathop{\mathbb{E}}_{q_\phi(\cx{0:T} | \ww)} [L_T] & \rightarrow \mathbb{E}[||\mathop{\mathbb{E}}_{\cx{T}\sim q}[\cx{T}| \cx{0}] - 0 ||^2] = \mathbb{E}[||\hat{\mu}(\cx{T}; \cx{0})] ||^2] \\
    \mathop{\mathbb{E}}_{q_\phi(\cx{0:T} | \ww)} [L_{t-1}] & \rightarrow \mathbb{E}[|| \mathop{\mathbb{E}}_{\cx{t-1}\sim q}[\cx{t-1}| \cx{0}, \cx{t}] - \mathop{\mathbb{E}}_{\cx{t-1} \sim \ptheta}  [\cx{t-1} | \cx{t}] ||^2] =\mathbb{E}[ ||\hat{\mu}(\cx{t}, \cx{0})-\mu_\theta(\cx{t}, t) ||^2]\\
    \mathop{\mathbb{E}}_{q_\phi(\cx{0:T} | \ww)} [L_{0}] & \rightarrow \mathbb{E}[|| \mathop{\mathbb{E}}_{\cx{0}\sim\qphi} [\cx{0} \mid \ww] - \mathop{\mathbb{E}}_{\cx{0} \sim \ptheta} [\cx{0}  \mid \cx{1} ] ||^2] = \mathbb{E}[ ||\Emb(w)-\mu_\theta(\cx{1}, 1) ||^2]
\end{align*}

It's worth noting that the first term is constant if the noise schedule satisfies $\alphabar_T = 0$, which guarantees $\cx{T}$ is pure Gaussian noise. 
In contrast, if the noise schedule doesn't go all the way such that $\cx{T}$ is pure Gaussian noise, we need to include this regularization term to prevent the embedding from learning too large norms. Embedding with large norms is a degenerate solution, because it is impossible to  sample from $p(\cx{T})$ accurately, even though it makes all the other denoising transitions easily predictable. 

Combining these terms yield $\Lsimpleee$. 
\begin{align*}
    \mathcal{L}^\text{e2e}_\text{simple} (\ww) & =  \mathop{\mathbb{E}}_{\qphi(\cx{0:T} | \ww)}  \left[||\hat{\mu}(\cx{T}; \cx{0})||^2 + \sum_{t=2}^T [||\hat{\mu}(\cx{t}, \cx{0})-\mu_\theta(\cx{t}, t) ||^2] \right] \\
& ~~~~~~~~~~~ + \mathop{\mathbb{E}}_{\qphi(\cx{0:1} | \ww)} \left[|| \Emb(\ww) - \mu_\theta(\cx{1}, 1) ||^2  - \log \ptheta(\ww | \cx{0})\right].
\end{align*}

Intuitively, we learn a Transformer model that that takes as input $(\cx{t}, t) \in (\mathbb{R}^{nd}, \mathbb{R})$ and the goal is to predict the distribution of $\cx{t-1} \in \mathbb{R}^{nd}$. It's worth noting that this Transformer model is shared across all the diffusion steps $t = 1\dots T$. As we demonstrated in the derivation of $\Lsimpleee$,  the most natural thing is to directly parametrize the neural network to predict the mean of $\cx{t-1} \mid \cx{t}$, we call this  $\mu_\theta$-parametrization. 

There are other parametrizations that are equivalent to $\mu_\theta$-parametrization up to a scaling constant. For example in \cref{ssec:rounding}, we can train the Transformer model to directly predict $\cx{0}$ via $\xxtheta(\cx{t}, t)$, and use the tractable Gaussian posterior $q(\cx{t-1} \mid \cx{0}, \cx{t})$ to compute the mean of $\cx{t-1}$, which has a closed form solution, conditioned on predicted $\cx{0}$ and observed $\cx{t}$: $ \frac{\sqrt{\alphabar_{t-1}} \beta_t}{1-\alphabar_t}\cx{0} + \frac{\sqrt{\alpha_t} (1-\alphabar_{t-1})}{1-\alphabar_t} \cx{t}$.
\begin{align*}
& ||\hat{\mu}(\cx{t}, \cx{0})-\mu_\theta(\cx{t}, t) ||^2 \\
= &  ||(\frac{\sqrt{\alphabar_{t-1}} \beta_t}{1-\alphabar_t}\cx{0} + \frac{\sqrt{\alpha_t} (1-\alphabar_{t-1})}{1-\alphabar_t} \cx{t}) - (\frac{\sqrt{\alphabar_{t-1}} \beta_t}{1-\alphabar_t}\xxtheta(\cx{t}, t) + \frac{\sqrt{\alpha_t} (1-\alphabar_{t-1})}{1-\alphabar_t} \cx{t})||^2\\
= & ||\frac{\sqrt{\alphabar_{t-1}} \beta_t}{1-\alphabar_t}(\cx{0} - \xxtheta(\cx{t}, t))||^2\\
\propto &  ||\cx{0} - \xxtheta(\cx{t}, t)||^2
\end{align*}
These two parametrizations differ by a constant scaling, and we apply the $\cx{0}$-parametrization to all terms in  $\Lsimpleee$ to reduce rounding errors as discussed in \cref{ssec:rounding}: 
\begin{align*}
    \mathcal{L}^\text{e2e}_{\cx{0}\text{-simple}} (\ww) & =  \mathop{\mathbb{E}}_{\qphi(\cx{0:T} | \ww)}  \left[||\hat{\mu}(\cx{T}; \cx{0})||^2 + \sum_{t=2}^T [||\cx{0}-\xxtheta(\cx{t}, t) ||^2] \right] \\
& ~~~~~~~~~~~ + \mathop{\mathbb{E}}_{\qphi(\cx{0:1} | \ww)} \left[|| \Emb(\ww) - \xxtheta(\cx{1}, 1) ||^2  - \log \ptheta(\ww | \cx{0})\right].
\end{align*}

To generate samples from a Diffusion-LM with $\cx{0}$-parametrization,  at each diffusion step, the model estimates the $\cx{0}$ via $\xxtheta (\cx{t}, t)$ and then we sample $\cx{t-1}$ from $q(\cx{t-1} \mid \xxtheta (\cx{t}, t), \cx{t})$, which is fed as input to the next diffusion step.

\section{Log-Likelihood Models and Results}
\label{app:simplex}

To investigate Diffusion-LM's log-likelihood performance, we make several departures from the training procedure of \cref{sec:method}.
Ultimately the log-likelihood improvements described in this section did not translate into better generation quality in our experiments and therefore we focus on the original method in the rest of the paper.
Our likelihood models are trained as follows:
\begin{itemize}
    \item Instead of training a diffusion model on sequences of low-dimensional token embeddings, we train a model directly sequences of on one-hot token vectors.
    \item Following the setup of \citet{kingma2021variational}, we train a continuous-time diffusion model against the log-likelihood bound and learn the noise schedule simultaneously with the rest of the model to minimize the loss variance.
    \item Because our model predicts sequences of one-hot vectors, we use a softmax nonlinearity at its output and replace all squared-error terms in the loss function with cross-entropy terms. This choice of surrogate loss led to better optimization, even though we evaluate against the original loss with squared-error terms.
    \item The model applies the following transformation to its inputs before any Transformer layers: $x := \mathrm{softmax}(\alpha(t) x + \beta(t))$ where $\alpha(t) \in \mathbb{R}$ and $\beta(t) \in \mathbb{R}^v$ are learned functions of the diffusion timestep $t$ parameterized by MLPs ($v$ is the vocabulary size).
    \item At inference time, we omit the rounding procedure in \cref{ssec:rounding}.
\end{itemize}
For exact model architecture and training hyperparameter details, please refer to our released code.

We train these diffusion models, as well as baseline autoregressive Transformers, on E2E and ROCStories and report log-likelihoods in \cref{tab:ppl}.
We train two sizes of Transformers: ``small'' models with roughly 100M parameters and ``medium'' models with roughly 300M parameters.
Both E2E and ROCstories are small enough datasets that all of our models reach their minimum test loss early in training (and overfit after that).
To additionally compare model performance in a large-dataset regime, we also present ``ROCStories (+GPT-J)'' experiments in which we generate 8M examples of synthetic ROCStories training data by finetuning GPT-J \citep{gpt-j} on the original ROCStories data, pretrain our models on the synthetic dataset, and then finetune and evaluate them on the original ROCStories data.

\begin{table*}
    \centering
\begin{tabular}{lcc|cc|cc|cc|cc}
\toprule
    Dataset                & Small AR    & Small Diffusion     & Medium Diffusion\\
    E2E                    & 1.77    & 2.28 & -\\
    ROCStories             & 3.05    & 3.88 & -\\
    ROCStories (+GPT-J)    & 2.41    & 3.59 & 3.10\\

\bottomrule
\end{tabular}
\caption{\label{tab:ppl} Log-likelihood results (nats per token)}
\end{table*}

\section{Qualitative Examples}
\label{app:qualitative}

We show randomly sampled outputs of Diffusion-LM  both for unconditional generation and for the $5$ control tasks. \cref{app:qualitative-unc} shows the unconditional generation results.
\cref{app:qualitative-span}, \cref{app:qualitative-pos}, \cref{app:qualitative-content}, and \cref{app:qualitative-syntax} show the qualitative samples from span control, POS control, semantic content control, and syntax tree control, respectively.
\cref{app:qualitative-length} shows the results of length control.

\section{Additional Ablation Studies}
\label{app:ablation}
\begin{figure}

 \vspace{-10pt}
    \centering
    \includegraphics[width=0.8\textwidth]{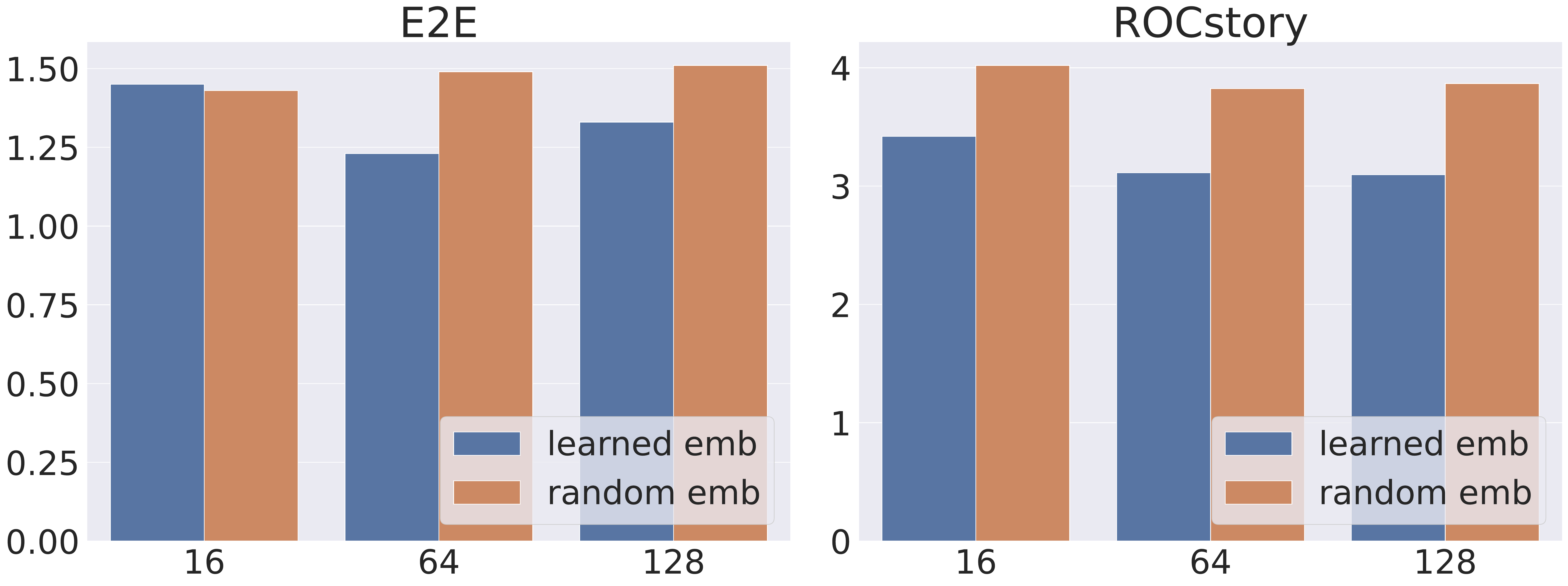}
    \includegraphics[width=0.8\textwidth]{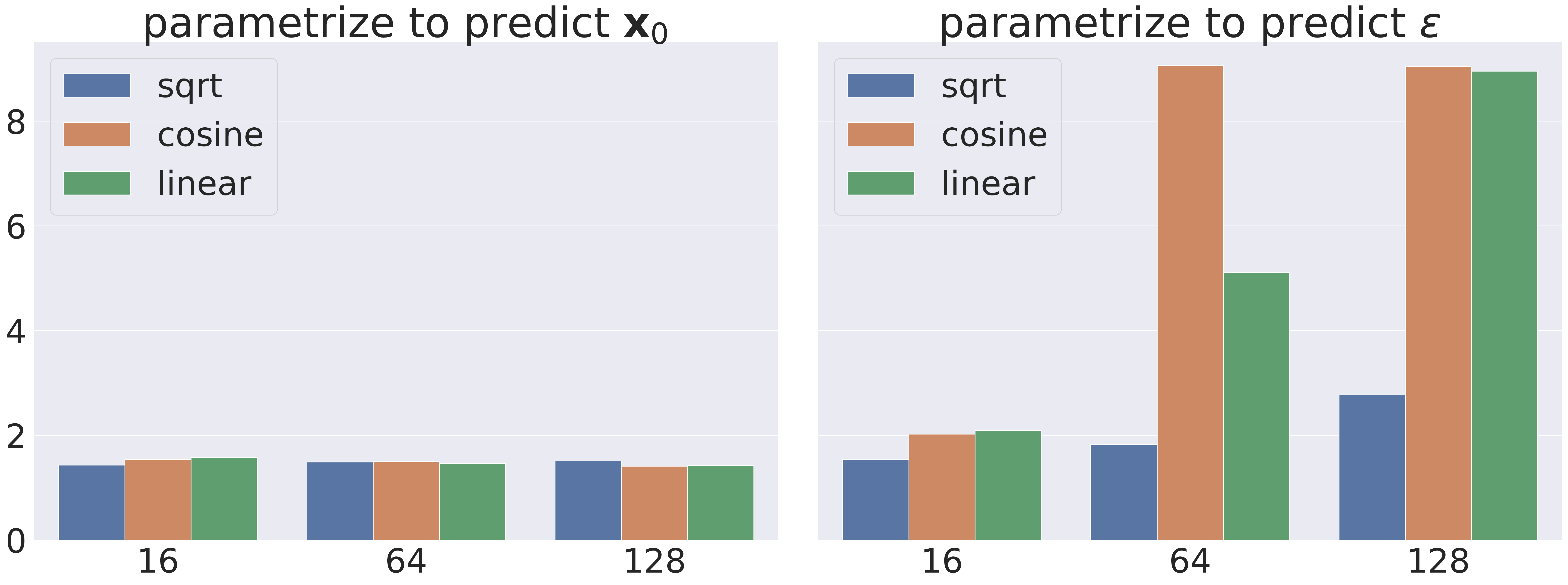}
    \includegraphics[width=0.4\textwidth]{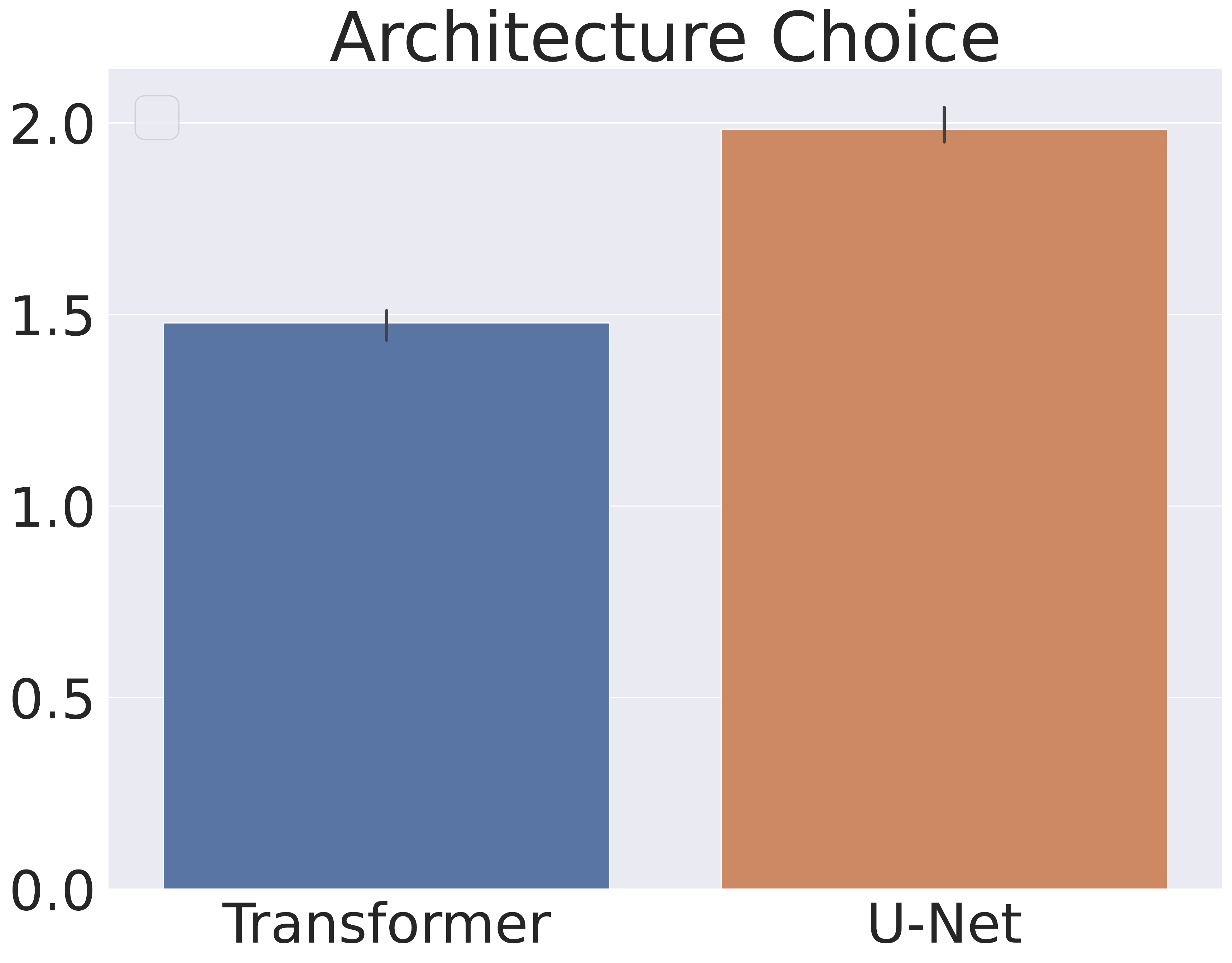}
\caption{\label{app:tab:ablation} Additional ablation results. The first row shows Diffusion-LM with trainable embeddings outperform random embeddings on both datasets (\cref{ssec:embeddings}). The second row demonstrates that \sqrtt schedule attains consistently good and stable performance across all dimension and parametrization choices. The last row shows that Transformer architecture outperforms U-Net architecture for language modeling. 
}
\end{figure}

In addition to the 2 ablation studies in \cref{sec:ablation}, we provide more ablation results in \cref{app:tab:ablation} about architecture choices and noise schedule. 

\textbf{Learned v.s. Random Embeddings (\cref{ssec:embeddings}).} 
Learned embeddings outperform random embeddings on both ROCStories and the E2E dataset by xx percent and xx percent respectively, as shown in the first row of \cref{app:tab:ablation}.

\textbf{Noise Schedule (\cref{app:noise}).}
We compare the \sqrtt schedule with \cosine \cite{nichol2021improved} and \linear \cite{ho2020denoising} schedules proposed for image modeling. The middle row of  \cref{app:tab:ablation} demonstrates that \sqrtt schedule attains consistently good and stable performance across all dimension and parametrization choices. While the \sqrtt schedule is less important with $\cx{0}$-parametrization, we see that it provides a substantially more robust noise schedule under alternative parametrizations such as $\epsilon$.

\paragraph{Transformer v.s. U-Net.}
The U-Net architecture in \citet{ho2020denoising} utilizes 2D-convolutional layers, and we imitate all the model architectures except changing 2D-conv to 1D-conv which is suitable for text data. 
\cref{app:tab:ablation} (last row) shows that the Transformer architecture outperforms U-Net.

\section{Societal Impacts}
On the one hand, having strong controllability in language models will help with mitigating toxicity, making the language models more reliable to deploy. Additionally, we can also control the model to be more truthful, reducing the inaccurate information generated by the language model by carefully controlling it to be truthful.   
On the other hand, however, one could also imagine more powerful targeted disinformation (e.g., narrative wedging) derived from the fine-grained controllability.  

Towards this end, it might be worth considering generation methods that can watermark the generated outputs without affecting its fluency, and this type of watermark could also be framed as a  controllable generation problem, with distinguish-ability and fluency as the constraints. 

\FloatBarrier

\begin{table*}
    \centering
    \resizebox{1\linewidth}{!}{
    \begin{tabular}{lp{13cm}}
\toprule
\parbox[t]{2mm}{\multirow{5}{*}{\rotatebox[origin=c]{90}{ROCStories+Aug}}} & Matt was at the store . He was looking at a new toothbrush . He found the perfect one . When he got home , he bought it . It was bright and he loved it . \\\\
& I and my friend were hungry . We were looking for some to eat . We went to the grocery store . We bought some snacks . We decided to pick up some snacks . \\\\
& I was at the store . I had no money to buy milk . I decided to use the restroom . I went to the register . I was late to work . \\\\
& The man wanted to lose weight . He did n't know how to do weight . He decided to start walking . He ate healthy and ate less . He lost ten pounds in three months . \\\\
& I went to the aquarium . I wanted to feed something . I ordered a fish . When it arrived I had to find something . I was disappointed .\\
\toprule
\toprule
\parbox[t]{2mm}{\multirow{5}{*}{\rotatebox[origin=c]{90}{ROCStories}}} & Tom was planning a trip to California . He had fun in the new apartment . He was driving , until it began to rain . Unfortunately , he was soaked . Tom stayed in the rain at the beach . \\\\
& Carrie wanted a new dress . She did not have enough money . She went to the bank to get one , but saw the missed . Finally , she decided to call her mom . She could not wait to see her new dress . \\\\
& Tina went to her first football game . She was excited about it . When she got into the car she realized she forgot her hand . She ended up getting too late . Tina had to start crying . \\\\
& Michael was at the park . Suddenly he found a stray cat . He decided to keep the cat . He went to his parents and demanded a leg . His parents gave him medicine to get it safe . \\\\
& Tim was eating out with friends . They were out of service . Tim decided to have a pizza sandwich . Tim searched for several hours . He was able to find it within minutes . \\
\toprule
\toprule
\parbox[t]{2mm}{\multirow{5}{*}{\rotatebox[origin=l]{90}{E2E}}} 
& The Waterman is an expensive pub that serves Japanese food . It is located in Riverside and has a low customer rating . \\\\
& A high priced pub in the city centre is The Olive Grove . It is a family friendly pub serving French food . \\\\
& The Rice Boat offers moderate priced Chinese food with a customer rating of 3 out of 5 . It is near Express by Holiday Inn . \\\\
& There is a fast food restaurant , The Phoenix , in the city centre . It has a price range of more than \u00a3 30 and the customer ratings are low .\\\\
& The Mill is a coffee shop based in the city centre area near to The Sorrento . It is in the high price range and serves Indian food . \\
\bottomrule
    \end{tabular}}
\caption{\label{app:qualitative-unc} Randomly sampled examples generated by unconditional sampling Diffusion-LM trained on 3 datasets. ROCStories+Aug denotes ROCStories with data augmentation. It's generated by first fine-tuning GPT-j on the ROCStories dataset and then sample the large GPT-j model to generate 1M stories. } 
\end{table*}

\begin{table*}
    \centering
    \resizebox{0.95\linewidth}{!}{
    \begin{tabular}{lp{13cm}}
\toprule
target span & [3, 5, PP] \\ 
\midrule
FUDGE &  UNK the UNK \greenspan{for Italian food} , The Eagle coffee shop is near Burger King in the riverside area . The Eagle has a customer rating of 5 out of 5 , and isn ' t family - friendly . The Eagle has a cheap price range . \\
Diffusion-LM &   The Plough , \greenspan{near Café Rouge} , is a high priced fast food pub . \\
FT &  Along the riverside \greenspan{near Café Rouge} is The Golden Curry . It serves Italian food in a family - friendly environment . It has a low customer rating . \\
\toprule
target span & [10, 12, PP]\\
\midrule
FUDGE &  Blue Spice is a high price range Fast food located \greenspan{in city centre} .\\
Diffusion-LM &   The Phoenix is a high priced food restaurant , located \greenspan{near the river} .\\
FT &  The Punter is a family restaurant with low prices and \errorspan{delicious sushi ,} located near the Café Sicilia \\
\toprule
target span & [9, 14, S]\\
\midrule
FUDGE & Zizzi pub serves Italian food for adults only . \errorspan{It has been rated average by} customers . \\
Diffusion-LM &  There is a Chinese restaurant called The Eagle , \greenspan{it has an average customer rating} . \\
FT & On the riverside area are located Alimentum , has \errorspan{a very good French food for} adults and kids , UNK price range are over 20 to 25 £ . \\
\toprule
target span & [4, 16, VP]\\
\midrule
FUDGE & The Cambridge Blue pub \errorspan{is near the Café Brazil and offers a high price range for their} French food . \\
Diffusion-LM & On the Ranch there \greenspan{is a children friendly pub called The Cricketers with an average customer rating} . \\
FT &  The Travellers Rest Beefeater \greenspan{is an average rated restaurant located in the riverside area near Café Adriatic} . Their price range is less than £ 20 . \\
\toprule
target span & [0, 2, NP]\\
\midrule
FUDGE & \greenspan{The Golden Palace} is a cheap , 5 - star coffee shop , located on the river in the north of the city centre . \\
Diffusion-LM & \greenspan{The Olive Grove} is a pub that provides Indian food in the high price range . It is in the city centre . \\
FT &  \greenspan{The Golden Curry} is located in city centre near Café Rouge which provides English food . Its customer rating is average and is not family - friendly . \\
\toprule
target span & [12, 13, NP]\\
\midrule
FUDGE & The Waterman is a family friendly place with a good rating .\errorspan{ [missing span]}\\
Diffusion-LM &   The Vaults is a high priced , family friendly restaurant that serves \greenspan{Italian food} . \\
FT &  Strada is a restaurant which costs less than £ 20 , but \errorspan{is not} family - friendly and has an average rating . \\
 
\bottomrule
\end{tabular}}
\caption{\label{app:qualitative-span}  Qualitative output of the syntax span control tasks. The target span ($i,j,$label) means the span from position $i$ to position $j$ should be a constituent with a specific label: S is sentence, NP is noun phrase, VP is verb phrase, PP is prepositional phrase, etc.  We color \errorspan{failed spans red} and \greenspan{correct spans green}. }
\end{table*}

\begin{table*}
    \centering
    \resizebox{0.95\linewidth}{!}{
    \begin{tabular}{lp{13cm}}
\toprule
target POS &  PROPN AUX DET ADJ NOUN NOUN VERB ADP DET NOUN ADP DET NOUN PUNCT\\
\midrule
FUDGE &  Aromi is a non family - friendly fast food coffee shop in the riverside area with a low Customer Rating . \\
Diffusion-LM &   Fitzbillies is a cheap coffee shop located on the outskirts of the city . \\
FT &  Aromi is a fast food pub located at the centre of the city. 
\\
\toprule
target POS &  PROPN AUX DET NOUN VERB NOUN ADJ NOUN PUNCT PRON NOUN NOUN AUX ADJ\\
\midrule
FUDGE &  Cocum is a family - friendly coffee shop , that has a low price range and a low customer rating . \\
Diffusion-LM &   Zizzi is a pub providing restaurant Chinese food . Its customer rating is low \\
FT &  Zizzi is a pub providing kids friendly services. Its customer rating is average 
\\
\toprule
target POS &  DET NOUN PUNCT PROPN VERB ADJ CCONJ ADJ NOUN CCONJ AUX VERB ADP DET PROPN ADJ PROPN PUNCT\\
\midrule
FUDGE &  A child - friendly coffee shop , Cocum , offers fast food at an average price range of £ 20 - 25 . \\
Diffusion-LM &   The Waterman - friendly serves UNK and fast food and is located near the Crown Plaza Hotel . \\
FT &  The wine - Strada serves fast and cheap food and is located near the Rainbow Vegetarian Café. 
\\
\toprule
target POS &  DET PROPN PROPN VERB ADJ NOUN ADP NOUN ADP SYM NUM PUNCT NOUN NOUN AUX ADJ PUNCT DET PROPN PROPN AUX VERB ADP DET PROPN CCONJ PROPN ADP PROPN PROPN PUNCT ADJ PUNCT DET NOUN PART AUX VERB PUNCT\\
\midrule
FUDGE &  The Midsummer House offers cheap English food near All Bar One . Rated 5 out of 5 . \\
Diffusion-LM &   The Rice Boat provides Chinese food in £ 20 - 25 . Price range is high . The Rice Boat is located near the Express by Holiday Inn and is kids friendly . The customer rating is high . \\
FT &  The Rice Boat welcomes Japanese food with prices under £ 20. Customer ratings are low. The Rice Boat is located near the Express by Holiday Inn. Convenient. No children's are allowed. 
\\
\toprule
target POS &  PROPN PROPN AUX DET ADJ NOUN NOUN ADP DET NOUN NOUN ADP DET PROPN PUNCT PRON AUX NOUN PUNCT ADJ PUNCT\\
\midrule
FUDGE &  Loch Fyne is a Japanese restaurant with a moderate price range and kid - friendly atmosphere . \\
Diffusion-LM &   Browns Cambridge is an Italian restaurant shop in the city centre near The Sorrento . It is family - friendly . \\
FT &  Browns Cambridge is a cheap coffee shop in the riverside area near The Sorrento, that is family - friendly. 
\\
\toprule
target POS &  PROPN VERB DET ADJ NOUN NOUN PROPN PUNCT PRON AUX ADJ VERB CCONJ VERB NOUN SCONJ VERB ADJ NOUN PUNCT\\
\midrule
FUDGE &  Fitzbillies coffee shop has a high price range , children friendly service and serves Japanese food in riverside with high customer rating . \\
Diffusion-LM &   There has a high customer rating . It is kid friendly called The Golden Curry and serves Indian food . \\
FT &  Customers give the French coffee shop Fitzbillies ; it is average rated and offers families where serving light meals. 
\\
\toprule
target POS &  DET NUM NUM VERB ADJ NOUN\\
\midrule
FUDGE &  The Twenty Two serves Fast food and is kids friendly . \\
Diffusion-LM &  The Twenty Two provides Chinese food \\
FT &  The Twenty Two provides Indian food 
\\
\toprule
target POS &  ADV NOUN ADV ADP PROPN PROPN PUNCT DET PROPN NOUN NOUN VERB ADJ NOUN NOUN CCONJ AUX PART VERB NOUN NOUN PUNCT\\
\midrule
FUDGE &  UNK your whole family to The Wrestlers , the best UNK the UNK UNK at the river \\
Diffusion-LM &   Located in riverside near The Sorrento , Browns Cambridge coffee shop serves Japanese food , and is not family - friendly . \\
FT &  Even adults only at Loch Fyne, The Rice Boat coffee shop has moderate price range and does not cater kids age. 
\\
\toprule
target POS &  DET PROPN AUX DET NUM NOUN NOUN NOUN VERB ADP DET PROPN PROPN PUNCT\\
\midrule
FUDGE &  The Eagle is a 3 star coffee shop located near Burger King , north of the City centre that provides low - cost fast food . \\
Diffusion-LM &   The Cricketers is a five star coffee shop located near The Portland Arms . \\
FT &  The Vaults is a one star coffee shop located near the Café Brazil. 
\\

\bottomrule
\end{tabular}}
\caption{\label{app:qualitative-pos}  Qualitative output of the POS control tasks. The target POS is the sequence of gold parts-of-speech tags the generated texts should match.  }
\end{table*}

\begin{table*}
    \centering
    \resizebox{0.95\linewidth}{!}{
    \begin{tabular}{lp{13cm}}
\toprule
target length &  7 \\
\midrule
FUDGE &  Wildwood is a cheap Japanese pub . \errorspan{Low rating .} \\
Diffusion-LM &   The Twenty Two serves Indian food . \\
FT &  The Mill is an Indian restaurant . \\
\toprule
target length &  12 \\
\midrule
FUDGE &  The Phoenix is an average Japanese restaurant that is in the City \errorspan{Centre . }\\
Diffusion-LM &   The Twenty Two serves Chinese food and is not family friendly . \\
FT &  Green Man is an average priced restaurant located near All Bar One \\
\toprule
target length &  17 \\
\midrule
FUDGE &  Fitzbillies is an expensive Italian coffee shop in the city centre . It is not child friendly\errorspan{ . .} \\
Diffusion-LM &  The Twenty Two serves Indian food in the city centre . It is not family friendly . \\
FT &  For low - priced food and a family - friendly atmosphere, visit Fitzbillies near Express by \errorspan{ Holiday Inn }
\\
\toprule
target length &  22 \\
\midrule
FUDGE &  The Golden Curry is an English food restaurant located near the Café Rouge in the Riverside area . The customer rating is \errorspan{average . Children are welcome .} \\
Diffusion-LM &   Strada is a fast food pub located near Yippee Noodle Bar and has a customer rating of 3 out of 5 . \\
FT &  There is an Italian kid friendly restaurant in the riverside area near The Sorrento named Browns Cambridge in the riverside area . \\
\toprule
target length &  27 \\
\midrule
FUDGE &  The Olive Grove is an expensive , children friendly , Fast food restaurant in the city centre . \errorspan{[missing 9 words]}\\
Diffusion-LM &   The Eagle is a family friendly coffee shop in the city centre near Burger King . It serves Italian food and has a low customer rating . \\
FT &  A pub in the city centre near Yippee Noodle Bar is named Strada. It serves French food and has a customer rating of 3 out of \errorspan{ 5 }
\\
\toprule
target length &  32 \\
\midrule
FUDGE &  The Golden Curry is a Japanese food restaurant with a high customer Rating , kid friendly and located along the riverside near Café Rouge . \errorspan{[missing 7 words]}\\
Diffusion-LM &   There is a family - friendly coffee shop in the city centre , it is called Zizzi . It is cheap and has a customer rating of 5 out of 5 . \\
FT &  In the city centre is a kids friendly place called Green Man. It has Japanese food and is near All Bar One. It has a price range of £ 20 \errorspan{ - 25 }
\\
\toprule
target length &  37 \\
\midrule
FUDGE &  There is a coffee shop called Fitzbillies which offers French food at cheap prices . It is not family - friendly and has a customer rating of 5 out of 5 . It is in riverside . \\
Diffusion-LM &   The Rice Boat provides Indian food in the moderate price range . It is located in the city centre . It is near Express by Holiday Inn . Its customer rating is 3 out of 5 . \\
FT &  For a family friendly coffee shop that serves Italian food, with a customer rating of 5 out of 5 and a cheap price range, try The Eagle. It is located in the riverside area \errorspan{. }
\\
\bottomrule
\end{tabular}}
\caption{\label{app:qualitative-length}  Qualitative output of the length control tasks, where all the generated texts tried to exactly match the target length. We mark the words exceeding the target length \errorspan{red}. }
\end{table*}

\begin{table*}
    \centering
    \resizebox{0.95\linewidth}{!}{
    \begin{tabular}{lp{13cm}}
\toprule
\toprule
target semantic content &  name : Bibimbap House \\
\midrule
FUDGE &  Clare Hall , the \greenspan{Bibimbap House} , serves high end Japanese food in the city centre . \\
Diffusion-LM &   \greenspan{Bibimbap House} in riverside near Clare Hall has a cheap price range . \\
FT &  By Clare Hall is \greenspan{Bibimbap House} which serves expensive noodles. 
\\
\toprule
target semantic content &  name : Travellers Rest Beefeater \\
\midrule
FUDGE &  \errorspan{Clowns} near Clare Hall in riverside is a French coffee shop rated 5 out of 5 \\
Diffusion-LM &   \errorspan{Green Man} is an Italian pub located in the city centre near Café UNK . \\
FT &  \greenspan{Travellers Rest Beefeater} is a reasonably priced restaurant that is family friendly. 
\\
\toprule
target semantic content &  Type : coffee shop \\
\midrule
FUDGE &  Wildwood is a \greenspan{coffee shop} located near Ranch . It is expensive and highly UNK . \\
Diffusion-LM &   The Punter is a high priced \greenspan{coffee shop} located near Café Sicilia that serves Japanese food . It is not family - friendly and has a customer rating of 3 out of 5 . \\
FT &  Located in the riverside area is the \greenspan{coffee shop} Fitzbillies. It has Indian food in the price Range of less than £ 20 and a low customer Rating. It is not family Friendly. 
\\
\toprule
target semantic content &  customer rating : low \\
\midrule
FUDGE &  The Waterman is a fast food restaurant that is family - friendly near the city centre . \errorspan{[missing content]}\\
Diffusion-LM &   The Rice Boat restaurant has a \greenspan{low customer rating} and is located in riverside . It serves Italian food , and is not family - friendly . \\
FT &  The Eagle is \greenspan{low customer rating} coffee shop with Italian food in riverside near Burger King. Its price range is less than £ 20 and is family - friendly. 
\\
\toprule
target semantic content &  near : The Sorrento \\
\midrule
FUDGE &  Browns Cambridge provides Indian food in the less than £ 20 price range . Its customer rating is low . \errorspan{[missing content]}\\
Diffusion-LM &   \greenspan{Near The Sorrento} on the riverside is a pub named Taste of Cambridge that serves Japanese food . \\
FT &  Browns Cambridge sells Italian food and is also a coffee shop. It has an average customer rating. It is located in the riverside area \errorspan{near Crowne Plaza Hotel} and yes it is child friendly. 
\\
\toprule
target semantic content &  food : Italian \\
\midrule
FUDGE &  A non family - friendly \greenspan{Italian} pub is Zizzi . It has an average customer rating .\\
Diffusion-LM &  Loch Fyne is \greenspan{Italian} Japanese restaurant that is kid friendly . \\
FT &   situated near All Bar One is a child friendly \greenspan{Italian} eatery called Green Man costing more than £ 30 is a restaurant near the riverside 
\\
\toprule
target semantic content &  price : high \\
\midrule
FUDGE &  The Vaults is a \greenspan{high priced} Italian Pub with a customer rating of 3 out of 5 near Café Adriatic \\
Diffusion-LM &   The Punter is a French restaurant with a \greenspan{high price} range . \\
FT &  A fast food coffee shop that is not kid friendly is called Cocum. It is \greenspan{expensive} and gets average ratings. 
\\
\end{tabular}}
\caption{\label{app:qualitative-content}  Qualitative output of the semantic content control task. We mark the compliant spans as green, and the spans that violates the control target as red. }
\end{table*}

\end{document}